\documentclass{article}

\usepackage{iclr2027_conference,times}
\usepackage{url}
\usepackage{booktabs}
\usepackage{graphicx}
\usepackage{amsmath,amssymb,amsthm}
\usepackage{microtype}
\usepackage{xcolor}
\usepackage{multirow}
\usepackage{tabularx}
\usepackage{array}
\usepackage{placeins}
\usepackage{float}
\usepackage{hyperref}

\hypersetup{
  pdftitle={What Do Tabular Foundation Models Compute In Context? In-Situ Representation Refinement through Attention-Gated Updates},
  pdfauthor={Tian Zhou, Beverly Jin, Linxiao Yang, Xue Wang, Wenwei Wang, Bingqing Peng, Mengni Ye, Jinjie Gu, Liang Sun},
  colorlinks=true,
  linkcolor=blue!45!black,
  citecolor=blue!45!black,
  urlcolor=blue!45!black
}

\newcolumntype{Y}{>{\raggedright\arraybackslash}X}
\newcommand{\modelname}{\textsc{RefineICL}}

\newtheorem{proposition}{Proposition}

\title{\Large What Do Tabular Foundation Models Compute\\
In Context?\\
In-Situ Representation Refinement through\\
Attention-Gated Updates}

\author{
\normalfont
Tian Zhou\textsuperscript{1,*},
Beverly Jin\textsuperscript{2,*},
Linxiao Yang\textsuperscript{1},
Xue Wang\textsuperscript{2},\\
Wenwei Wang\textsuperscript{1},
Bingqing Peng\textsuperscript{1},
Mengni Ye\textsuperscript{2},
Jinjie Gu\textsuperscript{1},
Liang Sun\textsuperscript{1}\\[3pt]
\textsuperscript{1}Ant Group
\qquad
\textsuperscript{2}Independent Researcher\\
\textsuperscript{*}Equal contribution
}

\iclrfinalcopy

\begin{document}
\maketitle
\lhead{}

\begin{abstract}
A tabular foundation model must discover which distinctions matter for each new
table without updating its parameters. We develop \emph{in-situ representation
refinement}: support labels guide changes to the episode's representations,
improving the information available to later queries.
A regularized leave-one-out objective yields
a support correction and its query extension. The leading term separates
attention-based reading from state-dependent scaling, motivating \modelname{}: an
attention-gated, FFN-free contextual stack with selected low-rank feature
interaction and typed memory. A direct intervention tests the role of evolving
support states: removing one intermediate support update while preserving the
block's query output increases final query cross-entropy in all
\textbf{72 tested episodes}. \modelname-L24 reaches \textbf{0.93836 OVR-AUC}
and \textbf{0.87173 accuracy} on AMLB29. A benchmark-informed continuation reaches
\textbf{1644.8 Elo} on the 38-dataset TabArena snapshot,
\textbf{31.4 Elo above TabPFN-3} under the same
evaluation. It also improves all four reported metrics over TabPFN-v3 on both
TabZilla views. In a matched 100K-update depth grid, an expanded FFN gives no
consistent validation benefit and uses \textbf{60.2\% more peak inference memory}
at L8. These results connect learning within a forward pass to representation
refinement and show how this view guides a competitive, memory-efficient model.
\end{abstract}

\section{Introduction}

The same rows can pose different learning problems. On a single point cloud,
one labeling asks for a linear separator, another for an XOR rule, and another
for a radial boundary. A tabular foundation model must infer the current rule
from labeled support rows and predict unlabeled queries in one forward pass,
with its pretrained parameters unchanged
\citep{mueller2022pfn,hollmann2025tabpfn,qu2025tabicl}.
The features alone do not say which distinctions matter.
\emph{How does the model turn the labels of a new table into a useful
representation for prediction?}

Support labels can supervise the representations used for prediction. When
other rows predict a support row incorrectly, they expose distinctions the
current states fail to capture. We call learning from this signal
\emph{in-situ representation refinement}. The support states become part of
the computation: in our model, removing an intermediate support update harms
later query predictions even when that block's query output is preserved.

\begin{figure*}[!t]
\centering
\includegraphics[width=\textwidth]{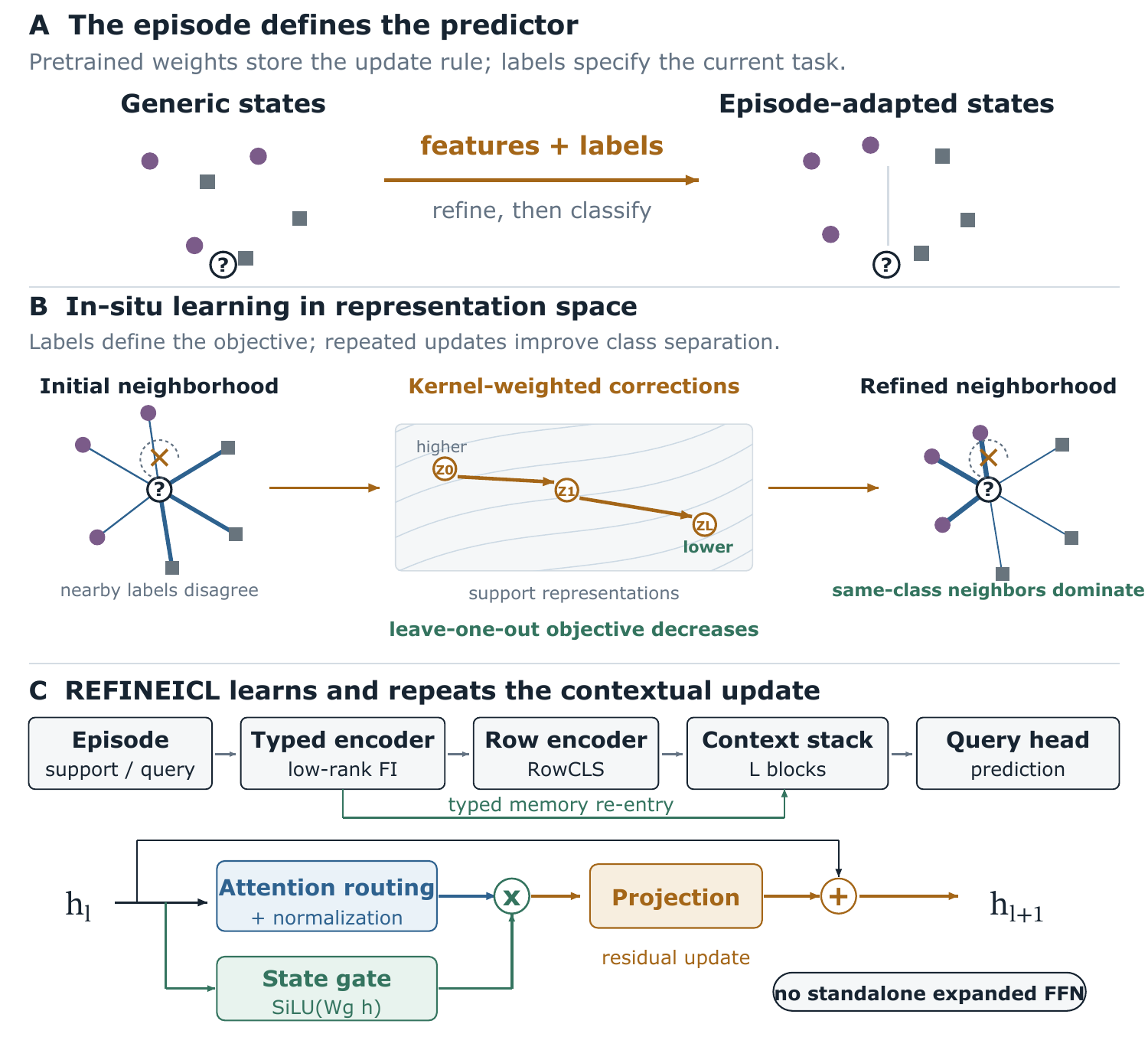}
\caption{\textbf{From the tabular ICL problem to \modelname{}.}
\textbf{A:} The labeled episode, rather than pretrained weights alone, determines the
predictor: support labels reshape generic states into an episode-adapted
representation before query prediction.
\textbf{B:} A schematic of the derived support-state correction: labels define
a leave-one-out objective, and kernel-weighted updates improve class geometry.
The depicted states are not measured network activations.
\textbf{C:} \modelname{} instantiates these roles with typed feature
encoding and selected low-rank feature interaction (FI), row compression, typed
memory re-entry, and repeated attention-gated residual updates without a
standalone expanded FFN. The complete architecture appears in Appendix
Figure~\ref{fig:refineicl-implementation-topology}.}
\label{fig:derivation-to-refineicl}
\end{figure*}

To make this account constructive, we ask what update would improve the
current episode. Prior optimization views show that Transformers can emulate
predictor updates in restricted settings
\citep{vonoswald2023transformers,akyurek2023learning}.
We derive updates to the representations themselves. Predicting each support
row from the others turns its known label into a test of the current geometry.
Regularized local optimization of this leave-one-out objective yields support
corrections, and a kernel extension transfers them to an unlabeled query.
Its leading term is
\begin{equation}
  u(z_q) \simeq \frac{1}{\alpha}\sum_i \kappa(z_q,z_i) f_i,
  \label{eq:intro-update}
\end{equation}
where $z_q$ and $z_i$ are query and support states, $\kappa$ measures similarity,
$f_i$ depends on support labels and current neighborhood error, and $\alpha>0$
controls regularization. The correction carried by a support row thus depends
on the other rows and changes as its representation changes. Normalizing the
similarities gives an attention readout but removes their total magnitude.
The leading-order correction retains that magnitude as a state-dependent
scale: where to read and how strongly to update are distinct parts of the
computation.

This separation motivates \modelname{}. Attention reads from evolving
support states, while a learned vector gate controls the update applied to
the receiving state, generalizing the scalar scale in the derivation.
Their product supplies a nonlinear residual update, repeated without a
standalone expanded feed-forward network (FFN). Low-rank feature interaction
and typed memory supply feature information for refinement.
A matched depth-by-FFN comparison tests the predictive benefit and memory
cost of adding expansion.

Our contributions connect this update structure to model design and prediction:

\begin{itemize}
  \item We formulate in-situ representation refinement through a leave-one-out
  support objective. Under local smoothness assumptions, we derive a support
  correction with a non-increase guarantee and extend it to unlabeled queries.
  Its leading-order form separates correction readout from update strength,
  providing a constructive basis for attention-gated residual blocks.
  \item We build \modelname{}, an attention-gated model with an FFN-free stack,
  selected low-rank feature interaction, and typed memory. L24 reaches
  \textbf{0.93836 OVR-AUC} on AMLB29 and, after benchmark-informed continuation,
  \textbf{1644.8 Elo} on TabArena (\textbf{31.4 above TabPFN-3} in the same
  evaluation). It improves all four reported metrics over TabPFN-v3 on both
  TabZilla views. In a matched depth grid, expansion gives no consistent
  validation gain and costs \textbf{60.2\% more peak inference memory} at L8.
  \item We show that updating the support representations contributes to
  subsequent prediction. Preserving a middle block's query output but skipping
  its support update increases final query cross-entropy in \textbf{72/72}
  episodes, by \textbf{0.0510} on average. Later blocks improve unlabeled-query
  nearest-prototype accuracy by \textbf{2.77 points}; paired interventions
  further test the roles of update assignment, representation dimensions,
  and distinct labeled examples.
\end{itemize}

\section{Related Work}

\paragraph{ICL as an optimization algorithm.}
Transformers can implement gradient descent, preconditioned updates, and
closed-form linear predictors in context
\citep{vonoswald2023transformers,akyurek2023learning}. These results establish an
important optimizer view of ICL, while subsequent work questions whether a
literal gradient-descent equivalence transfers to pretrained models outside
those controlled settings \citep{shen2024position}. This leaves a design
question: which states should a tabular model update as it reads a new task?
We study episode representations, using supervised neighborhood learning
\citep{goldberger2004nca,frosst2019soft}; the contribution is its regularized
support correction, extension to unlabeled queries, and use in model design.

\paragraph{Conditional interaction versus standalone FFNs.}
Transformer FFNs have been analyzed as parameterized key--value memories
\citep{geva2021ffn}. Recommendation models instead emphasize efficient explicit
feature interaction: DCN and DCN-V2 construct bounded-degree and low-rank feature
crosses \citep{wang2017dcn,wang2021dcnv2}, while HSTU fuses contextual aggregation
with multiplicative gating \citep{zhai2024hstu}. These operations offer a way
to make each update depend on the current table. \modelname{} connects them to
representation refinement: low-rank interactions prepare feature information,
and attention-gated blocks use it to update row states. We test this connection
through support labels, query state, and depth.

\paragraph{Tabular foundation models.}
Prior-data fitted networks amortize Bayesian-style prediction over synthetic
tasks \citep{mueller2022pfn}. TabPFNv2 established strong small-data performance
\citep{hollmann2025tabpfn}, and subsequent systems extend table scale, training
distributions, and test-time computation
\citep{grinsztajn2025tabpfn25,grinsztajn2026tabpfn3}. TabICL separates column
representation from row-level inference \citep{qu2025tabicl,qu2026tabiclv2};
TabFM scales alternating feature and row computation \citep{tabfm2026}; and
EXAONE-Tabular demonstrates that compact interleaved architectures remain
competitive \citep{eo2026exaone}. We study how a pretrained model uses labeled
rows to form a new predictor, using controlled experiments on internal states in
the medium-scale regime.

\paragraph{Evaluation.}
We retain each public benchmark's dataset list and aggregation rule
\citep{erickson2025tabarena,bischl2021openml,gijsbers2024amlb,
grinsztajn2022tree,mcelresh2023tabzilla,liu2024talent}.

\begin{figure}[!t]
\centering
\includegraphics[width=\linewidth]{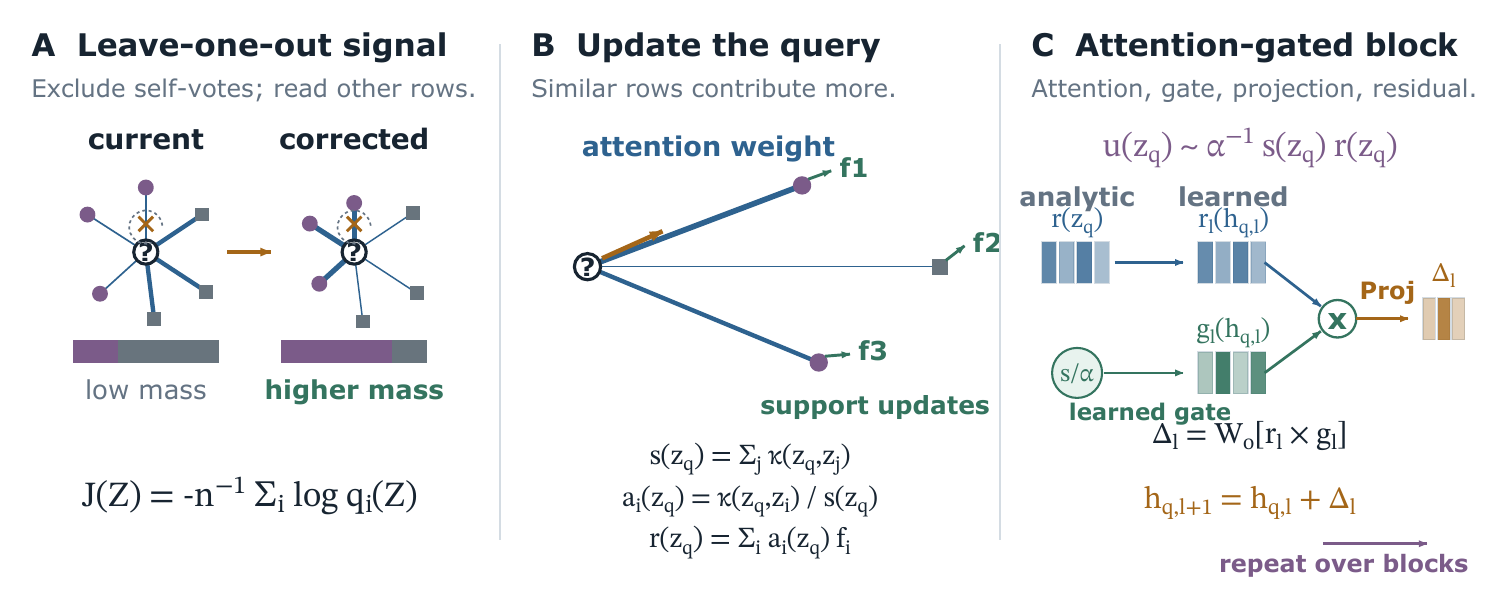}
\caption{\textbf{From a supervised objective to a learned contextual update.}
\textbf{A:} Excluding each row's own vote defines its correct-class neighbor
mass $q_i$ and the leave-one-out objective $\mathcal{J}$.
\textbf{B:} Kernel extension transfers support corrections to an unlabeled query;
normalizing the similarities gives the attention output $r(z_q)$, while the total
similarity $s(z_q)/\alpha$ sets the update size.
\textbf{C:} \modelname{} learns an attention output and uses a query-dependent
vector gate to scale it, then projects, adds, and repeats the update.}
\label{fig:refinement-derivation}
\end{figure}

\section{In-Situ Representation Refinement}
\label{sec:in-situ-theory}

The support rows provide a learning signal before any query answers are known:
predict each row from the others and use its label to identify errors. We derive
a representation correction, transfer it to queries, and use its form to guide
an attention-gated block (Figure~\ref{fig:refinement-derivation}).

An episode supplies labeled support rows
$\mathcal{S}=\{(x_i,y_i)\}_{i=1}^{n}$ and unlabeled query rows
$\mathcal{Q}=\{x_j\}_{j=n+1}^{n+m}$. To compute
$p_\theta(y_j\mid\mathcal{S},x_j)$ without updating $\theta$, we study how support
labels can refine the episode representation and transfer that refinement to
queries (Figure~\ref{fig:derivation-to-refineicl}B).

\subsection{What makes a representation useful for this episode?}

\paragraph{Labels define a local geometric target (Figure~\ref{fig:refinement-derivation}A).}
Predict each support row from the other labeled rows. Let
$Z=[z_1,\ldots,z_n]\in\mathbb{R}^{d\times n}$ be the current
support states, $y_i\in\{1,\ldots,C\}$, and
$\kappa(z_i,z_j)=\exp(\tau z_i^\top z_j)$ their similarity, where
$\tau>0$ is an inverse temperature. For $j\ne i$, define
\begin{equation}
 a_{ij}(Z)=\frac{\kappa(z_i,z_j)}{\sum_{k\ne i}\kappa(z_i,z_k)},
 \qquad
 q_i(Z)=\sum_{j\ne i}a_{ij}(Z)\mathbb{1}[y_j=y_i].
 \label{eq:soft-neighborhood}
\end{equation}
The correct-class neighbor mass $q_i$ excludes self-votes: only the other
support rows contribute to its prediction. With at least two support rows per class,
the following loss is finite:
\begin{equation}
 \mathcal{J}(Z)=-\frac{1}{n}\sum_i\log q_i(Z).
 \label{eq:loo-objective}
\end{equation}
This soft-neighbor objective \citep{goldberger2004nca,frosst2019soft} makes the
episode's labels a criterion for improving its states. We derive that
improvement before considering how a network can learn to produce it.

\paragraph{Support labels determine an update direction.}
The descent direction $F=-\nabla_Z\mathcal{J}(Z)=[f_1,\ldots,f_n]$ combines
support states according to similarity, label agreement, and neighborhood error.
Thus the same label calls for different corrections as its neighborhood
changes. We regularize this direction using support similarity, coupling the
corrections across rows.

\begin{proposition}[A support update that decreases the objective]
Let $K=[\kappa(z_i,z_j)]_{ij}+\epsilon I_n\succ0$ be the support Gram matrix
with a small diagonal term $\epsilon>0$, where $I_n$ is the $n\times n$ identity, and let
$\alpha>0$ be its regularization strength.
If $\mathcal{J}$ is $\beta$-smooth on a neighborhood containing the update
segment, where $\beta$ is the smoothness constant, the regularized local objective in
Appendix~\ref{sec:support-correction-details} has support-state solution
$U^*\in\mathbb{R}^{d\times n}$ satisfying
\begin{align}
 U^*&=FK(\beta K+\alpha I_n)^{-1},
 \label{eq:support-update}\\
 \mathcal{J}(Z+U^*)-\mathcal{J}(Z)
 &\le -\tfrac12\langle F,U^*\rangle\le0.
 \label{eq:descent-bound}
\end{align}
\end{proposition}
\subsection{How does the correction reach an unlabeled query?}

\paragraph{Similarity transfers the correction (Figure~\ref{fig:refinement-derivation}B).}
The remaining challenge is to update a row whose label is unknown. For query state $z_q$, define
$k_q=[\kappa(z_q,z_1),\ldots,\kappa(z_q,z_n)]^\top$. The smooth kernel extension
of the support correction is
\begin{equation}
 u(z_q)=F(\beta K+\alpha I_n)^{-1}k_q.
 \label{eq:query-update-exact}
\end{equation}
Its leading term when $\beta\lVert K\rVert_2/\alpha$ is small is
\begin{equation}
 u(z_q)\approx\frac{1}{\alpha}\sum_i
 \underbrace{\kappa(z_q,z_i)}_{\text{how much to read}}
 \underbrace{f_i}_{\text{what correction to carry}}.
 \label{eq:query-update-attention}
\end{equation}
The query borrows corrections from labeled rows according to similarity; its
own label is never needed. Appendix~\ref{sec:functional-extension} establishes
consistency with the support update.

The leading-order update can be written as follows. Let
\begin{equation}
 \begin{aligned}
 s(z_q)&=\sum_j\kappa(z_q,z_j), &
 a_i(z_q)&=\frac{\kappa(z_q,z_i)}{s(z_q)},\\
 r(z_q)&=\sum_i a_i(z_q)f_i, &
 u(z_q)&\approx \frac{s(z_q)}{\alpha}r(z_q).
 \end{aligned}
 \label{eq:query-update-factorized}
\end{equation}
The coefficients $a_i(z_q)$ are nonnegative and sum to one; equivalently, they
are softmax attention weights with logits $\log\kappa(z_q,z_i)$. Hence $r(z_q)$
has the form of attention over the support updates $f_i$. Normalization alone
loses the total similarity $s(z_q)$. Recovering the derived update requires the
query-dependent scale $s(z_q)/\alpha$. This gives a reason to separate
\emph{where and what to read} from \emph{how strongly to apply it}.

\subsection{How does this computation guide a neural block?}

\paragraph{Read, scale, and update (Figure~\ref{fig:refinement-derivation}C).}
The equations suggest two interacting branches. Attention reads a correction
from the evolving support states, while a state-dependent gate controls how
strongly the receiving state applies it. RefineICL uses a vector-valued SiLU
gate so that this scaling can differ across representation dimensions. Their
product is projected and added to the residual state:
\begin{equation}
 h_{\ell+1}=h_\ell+
 \underbrace{W_{o,\ell}\!\left[
 \underbrace{\operatorname{Norm}(A_\ell(\bar h_\ell))}_{\text{attention output}}
 \odot
 \underbrace{\operatorname{SiLU}(W_{g,\ell}\bar h_\ell)}_{\text{query-dependent gate}}
 \right]}_{\text{residual correction}}.
 \label{eq:context-update}
\end{equation}
The scalar scale motivates the learned vector gate; the block need not compute
the analytic correction exactly. Here $h_\ell$ collects the residual states at block $\ell$ and
$\bar h_\ell=\operatorname{LayerNorm}(h_\ell)$ is fed to both branches;
$A_\ell(\bar h_\ell)$ is the learned attention readout; $\operatorname{Norm}$ is its
normalization; $W_{g,\ell}$ and $W_{o,\ell}$ are the gate and output projections;
and $\odot$ is elementwise multiplication.

\paragraph{Three design consequences.}
\textbf{Update the support states:} the correction $f_i$ depends on the current
neighborhood, not just the label, motivating support updates before later query
reads. \textbf{Separate reading from scaling:} attention normalizes relative
similarities, while a gate controls the correction applied to the receiving state.
\textbf{Use the product as the nonlinear update:} this computation does not
require a separate expanded FFN. Section~\ref{sec:results} tests these choices
through support-update removal, query-specific attention/gate interventions,
and the performance--memory cost of FFN expansion, respectively.

\paragraph{What can make refinement more effective?}
\textbf{Depth} offers further updates; \textbf{width} gives them representation
dimensions; \textbf{context} supplies the labels that define the task.
Section~\ref{sec:three-resources} tests whether the learned model uses these
resources to improve its predictions.

\section{RefineICL: Attention-Gated Representation Refinement}
\label{sec:refineicl-method}

\subsection{Architecture}

\modelname{} learns to refine row states from the current labeled table.
Feature interactions and memory retain information through row compression
(Figure~\ref{fig:derivation-to-refineicl}C;
Appendix Figure~\ref{fig:refineicl-implementation-topology}).

\paragraph{Attention-gated contextual refinement.}
Every contextual block implements Equation~\ref{eq:context-update}:
$A_\ell$ selects and combines information from the support rows,
$g_\ell=\operatorname{SiLU}(W_{g,\ell}\bar h_\ell)$ scales each feature, and
$W_{o,\ell}$ adds the result to the residual state. Their product makes the
correction depend nonlinearly on context and the receiving state, without a
post-attention expanded FFN.

\paragraph{Selected low-rank feature interaction.}
Refinement needs row states that retain useful feature relations. Before
compression, a target-aware encoder represents numerical values, categorical
identity, and missingness separately. Two feature-graph rounds score
candidate feature pairs; only the 16 strongest neighbors of feature $i$ contribute to its update:
\begin{equation}
 m_i=\sum_{j\in\mathcal{N}_{16}(i)}p_{ij}W_o^{\mathrm{FI}}
 \left[(W_u^{\mathrm{FI}} h_i)\odot(W_v^{\mathrm{FI}} h_j)\right],
 \qquad h_i' = h_i+\gamma_{\mathrm{FI}} m_i.
 \label{eq:refineicl-low-rank-fi}
\end{equation}
Here $h_i$ is feature $i$'s state, $\mathcal{N}_{16}(i)$ contains its 16
highest-scoring partners, $p_{ij}$ is the normalized selected-edge weight,
$W_u^{\mathrm{FI}},W_v^{\mathrm{FI}},W_o^{\mathrm{FI}}$ are learned low-rank
projections, and $\gamma_{\mathrm{FI}}$ scales the residual update. All feature
pairs are scored, but messages are computed only on selected edges;
pre-compression and row-encoder interactions use ranks 32 and 64, respectively.

\paragraph{Row compression and typed-memory re-entry.}
A three-block row encoder and RowCLS tokens compress a variable feature set into
fixed-width row states. Later updates may need details lost in this summary. \modelname{} stores the
feature states after two interaction rounds before RowCLS and retrieves them with eight-head gated
reads at relative depths $1/3$ and $2/3$. Each read is initialized at zero and
limited to $0.25$ of the current row-state RMS. Later blocks can thus revisit
feature information without carrying all feature tokens through the stack.
Component tests find a larger effect for gating than for memory and individual
feature interactions (Appendix~\ref{sec:component-recovery-protocol}).

\subsection{Model sizes}

L12 and L24 use feature embedding dimension 256, with $(12,2048)$ and $(24,1024)$
contextual $(\text{depth},\text{width})$. The controlled FFN comparison
in Section~\ref{sec:accuracy-efficiency} holds depth and width fixed while toggling the expanded FFN.

\section{Experimental Design}

\paragraph{Public benchmarks.}
We evaluate L12 and L24 without dataset-specific updates. AMLB29, CC18,
Grinsztajn, and TabZilla use saved splits and four members; TabArena uses its
38-dataset, eight-member snapshot. Synthetic episode simulation and the training
policy follow TabICLv2 \citep{qu2026tabiclv2}. We train with distributed data
parallelism on 64 T-Head PPU-ZW810 accelerators (96 GiB each). Appendix~\ref{app:benchmarks}
gives the benchmark protocols; the code supplement provides exact recipes.

\paragraph{Mechanism tests.}
To connect prediction to refinement, fixed episodes test class separation,
representation dimensions, and support-set size. Query permutations change
which row receives an update while preserving the vectors. Paired bootstrap
intervals and controls are defined in Appendix~\ref{sec:mechanism-protocols}.

\paragraph{Controlled FFN comparison.}
Matched $d=96$ models train at L2/L4/L8 for 100K updates with a shared seed,
toggling FFN-2$\times$ in column, row, and contextual blocks together. We compare
held-out validation and, at L8, 54 boundary-rule conditions with 128--1,024
support rows and peak inference memory (Appendix~\ref{sec:depth-ffn-ablation}).

\paragraph{Component ablations.}
We disable one model component at inference on identical inputs; a matched recovery
study tests compensation (Appendix~\ref{sec:component-recovery-protocol}). Paired
comparisons measure how much each component affects the trained model.

\section{Results}\label{sec:results}

Prediction tests the practical value of this design. We then look inside the
forward pass: whether support states improve, whether later queries benefit,
and how changing labels or reassigning updates changes the predictor.

\subsection{Accuracy and efficiency of \modelname{}}\label{sec:accuracy-efficiency}

On real tables, L24 reaches 0.87173 accuracy and 0.93836 OVR-AUC on AMLB29,
exceeding L12 by 1.51 and 0.84 points and TabPFN-v3 by 0.30 and 0.22 points.
Benchmark-informed continuation reaches 1644.8 TabArena Elo, 31.4 above
TabPFN-3 in the same evaluation.

\begin{table}[!t]
\caption{\textbf{Classification under two separate evaluation protocols.}
\textbf{(A)} TabArena: 38 datasets, 78 methods; bold marks the best listed ICL
model below 200M parameters. \textbf{(B)} AMLB29: 29 tasks, ten folds,
four-member ICL. Panel A uses a benchmark-informed L24 continuation
(Appendix~\ref{app:benchmark-provenance}); HPO references are labeled.}
\label{tab:main-benchmarks-v2}
\centering
\footnotesize
\setlength{\tabcolsep}{2.8pt}
\textbf{(A) TabArena v0.1 snapshot}\par\vspace{2pt}
\begin{tabular}{lrrrr}
\toprule
Method & Elo $\uparrow$ & Mean rank $\downarrow$ & Win rate $\uparrow$ & Mean error $\downarrow$ \\
\midrule
\multicolumn{5}{l}{\textit{ICL models below 200M parameters}} \\
\textbf{\modelname-L24} & \textbf{1644.8} & \textbf{11.0789} & \textbf{0.86910} & \textbf{0.16238} \\
TA-TabPFN-3 (snapshot) & 1613.4 & 12.2368 & 0.85407 & 0.16751 \\
TabPFN-v2.6 (snapshot) & 1581.7 & 13.8684 & 0.83288 & 0.16675 \\
TabICLv2 (snapshot) & 1576.8 & 14.1316 & 0.82946 & 0.16895 \\
\addlinespace[2pt]
\multicolumn{5}{l}{\textit{AutoML and tuned ensemble references}} \\
AutoGluon 1.5 (extreme, 4h) & 1619.9 & 11.9211 & 0.85817 & 0.15470 \\
RealTabPFN-v2.5 (tuned + ens.) & 1566.4 & 14.7105 & 0.82194 & 0.16598 \\
\addlinespace[2pt]
\multicolumn{5}{l}{\textit{Large-model reference}} \\
TabFM (default; 1.639B parameters) & 1743.8 & 7.1316 & 0.92037 & 0.16203 \\
\bottomrule
\end{tabular}
\par\vspace{4pt}
\textbf{(B) AMLB29: matched ICL and published references}\par\vspace{2pt}
\begin{tabular}{llrrr}
\toprule
Model & Setting & Folds & Accuracy $\uparrow$ & OVR-AUC $\uparrow$ \\
\midrule
\modelname-L12 & 4 members & 290 & 0.85660 & 0.92999 \\
\textbf{\modelname-L24} & 4 members & 290 & \textbf{0.87173} & \textbf{0.93836} \\
TabPFN-v3 & 4 members & 290 & 0.86877 & 0.93614 \\
TabPFN-v2 (Nature default) & published; 4 members & 290 & -- & 0.93232 \\
TabPFN-v2 + PHE & published; HPO 4h & 290 & -- & 0.93319 \\
AutoGluon & published; HPO 4h & 290 & -- & 0.92564 \\
\bottomrule
\end{tabular}
\end{table}

Table~\ref{tab:benchmark-breadth-ffn}A extends the comparison to CC18,
Grinsztajn, and TabZilla. L24 improves AUC over TabPFN-v3 on 27/36 TabZilla
Canonical datasets and 68/102 Extension datasets (paired analysis in
Appendix~\ref{app:paired-benchmarks}). Its 159.5M parameters are 9.7\% of
TabFM's 1.639B (Appendix Table~\ref{tab:capacity-reference}).

\begin{table}[!t]
\caption{\textbf{Benchmark breadth and the cost of FFN expansion.}
\textbf{(A)} Four-member accuracy/AUC; bold marks the best displayed mean.
Overlapping views are not pooled. \textbf{(B)} FFN minus no-FFN at 100K updates,
$d=96$, one seed (Appendix Table~\ref{tab:depth-ffn-ablation}). Positive CE is
worse; positive accuracy is better. L8 memory is measured at 1,024 support rows
(Appendix~\ref{sec:ffn-context-memory-protocol}).}
\label{tab:benchmark-breadth-ffn}
\centering
\footnotesize
\setlength{\tabcolsep}{2.2pt}
\textbf{(A) Three additional benchmark families, six registered views}\par\vspace{2pt}
\begin{tabular*}{\linewidth}{@{\extracolsep{\fill}}lrrrrr@{}}
\toprule
Benchmark & $N$ & L12 & L24 & TabPFN-v3 & $\Delta$(L24$-$v3) \\
\midrule
OpenML-CC18 & 72 & 0.8663 / 0.9328 & 0.8819 / \textbf{0.9383} & \textbf{0.8860} / 0.9370 & $-0.41$ / $+0.13$ \\
Grinsztajn Current & 23 & 0.7811 / 0.8493 & \textbf{0.8176 / 0.8834} & 0.7992 / 0.8664 & $+1.85$ / $+1.71$ \\
\quad Categorical & 7 & 0.7354 / 0.8059 & \textbf{0.7967 / 0.8637} & 0.7644 / 0.8363 & $+3.23$ / $+2.73$ \\
\quad Numerical & 16 & 0.8010 / 0.8682 & \textbf{0.8268 / 0.8921} & 0.8144 / 0.8795 & $+1.24$ / $+1.26$ \\
TabZilla Canonical & 36 & 0.8520 / 0.9198 & \textbf{0.8676 / 0.9282} & 0.8615 / 0.9247 & $+0.61$ / $+0.35$ \\
TabZilla Extension & 102 & 0.8763 / 0.9189 & \textbf{0.8887 / 0.9277} & 0.8853 / 0.9227 & $+0.34$ / $+0.51$ \\
\bottomrule
\end{tabular*}
\par\vspace{4pt}
\textbf{(B) FFN-free gating avoids costly expansion}\par\vspace{2pt}
\setlength{\tabcolsep}{4.5pt}
\begin{tabular}{lrrl}
\toprule
Depth & $\Delta$ validation CE $\downarrow$ & $\Delta$ accuracy (pp) $\uparrow$ & Peak MiB, no-FFN $\rightarrow$ FFN \\
\midrule
L2 & $+0.00393$ & $+0.46$ & -- \\
L4 & $+0.02932$ & $-0.20$ & -- \\
L8 & $-0.01144$ & $+0.27$ & $202.97\rightarrow325.12$ (\textbf{+60.2\%}) \\
\bottomrule
\end{tabular}
\end{table}

Expansion offers an uneven return: whole-model FFNs worsen CE at L2/L4;
their L8 gain costs 60.2\% more peak memory. The 1,024-row boundary-rule gain is
1.11 accuracy points (Appendix~\ref{sec:ffn-context-memory-protocol}). Gated
refinement thus offers competitive prediction with lower memory cost.
Full results and recovery tests appear in Appendices~\ref{sec:depth-ffn-ablation} and
\ref{sec:component-recovery-protocol}.

\subsection{Depth, width, and evidence scale contextual refinement}
\label{sec:three-resources}

\begin{figure}[!t]
\centering
\includegraphics[width=\textwidth]{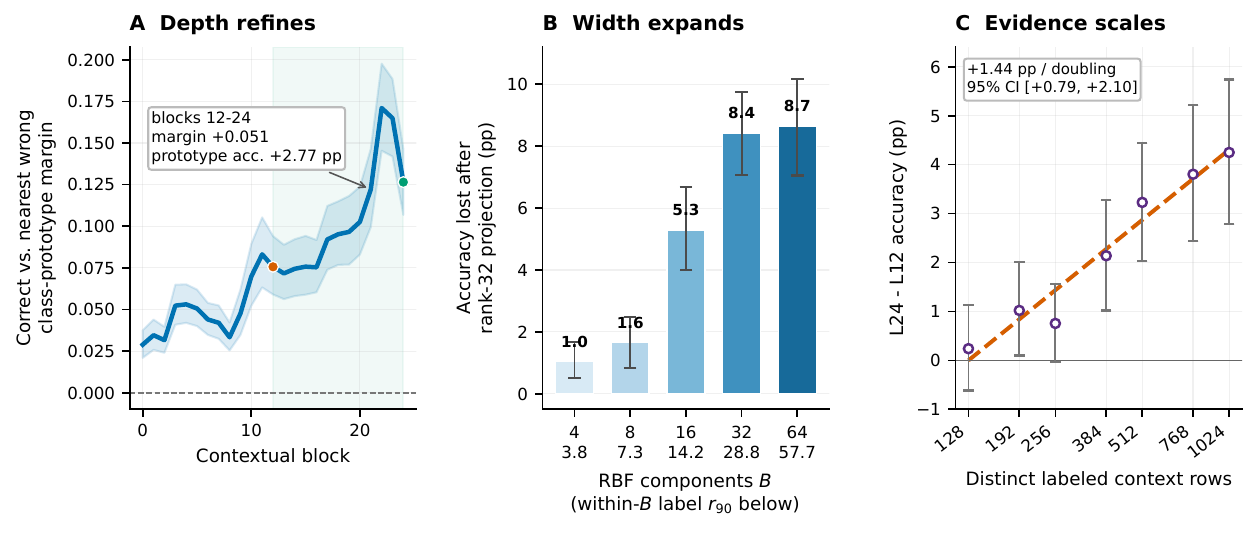}
\caption{\textbf{How depth, width, and labeled context support representation refinement.}
\textbf{A:} Class separation across blocks. \textbf{B:} Accuracy lost after
projecting the hidden state to rank 32 as the rule uses more RBF components. \textbf{C:} The paired
L24--L12 gap as distinct labeled rows increase. Intervals are 95\% bootstraps.}
\label{fig:three-resources-v2}
\label{fig:three-resources}
\end{figure}

A strong final prediction does not show where refinement helped.
Figure~\ref{fig:three-resources-v2} examines the states along the way, the
dimensions they use, and what additional labels contribute.

\paragraph{Depth improves predictions from the task representation.}
Useful refinement should make queries easier to classify from the states.
On 72 held-out RBF episodes, normalized support-class means serve as prototypes.
Blocks 12--24 increase the query correct-versus-wrong margin by 0.0509
(95\% CI [0.0403, 0.0616]) and nearest-prototype accuracy by 2.77 points,
without query labels as input.
Learned updates also reduce the leave-one-out support objective, while
row-permuted updates increase it (Appendix~\ref{sec:learned-support-update}).

\paragraph{Later queries benefit from updated support states.}
To test whether the support change matters, we preserve block 12's query output
but replace its support output with its input. Later blocks run normally.
Final query CE rises in all 72 episodes, by 0.0510 on average (95\% CI
[0.0417, 0.0608]); accuracy falls by 1.44 points. The support update therefore
helps subsequent prediction beyond the current query update
(Appendix~\ref{sec:learned-support-update}).

\paragraph{Width provides dimensions for task-specific updates.}
To test whether broader rules need more dimensions, we fix eight input
features and construct rules from the first $B$ support-kernel
eigenfunctions. Projecting the 1,024-dimensional state after block 4 to rank 32
costs 1.04 accuracy points at $B=4$ but 8.66 at $B=64$ (increase 7.62;
95\% CI [6.03, 9.05]). The dimensions removed matter more for broader rules.

\paragraph{Additional labeled rows widen the L24--L12 gap.}
L24 makes greater use of new labels. Across 72 matched cells with fixed queries,
increasing distinct support rows from 128 to 1,024 gives L24 23.86 accuracy
points versus L12's 19.85. Their gap grows from 0.24 to 4.25 points
(increase 4.01; 95\% CI [2.26, 5.79]).

\FloatBarrier
\subsection{Support labels and query-specific updates}

\begin{figure}[H]
\vspace{-3pt}
\centering
\includegraphics[width=0.76\textwidth]{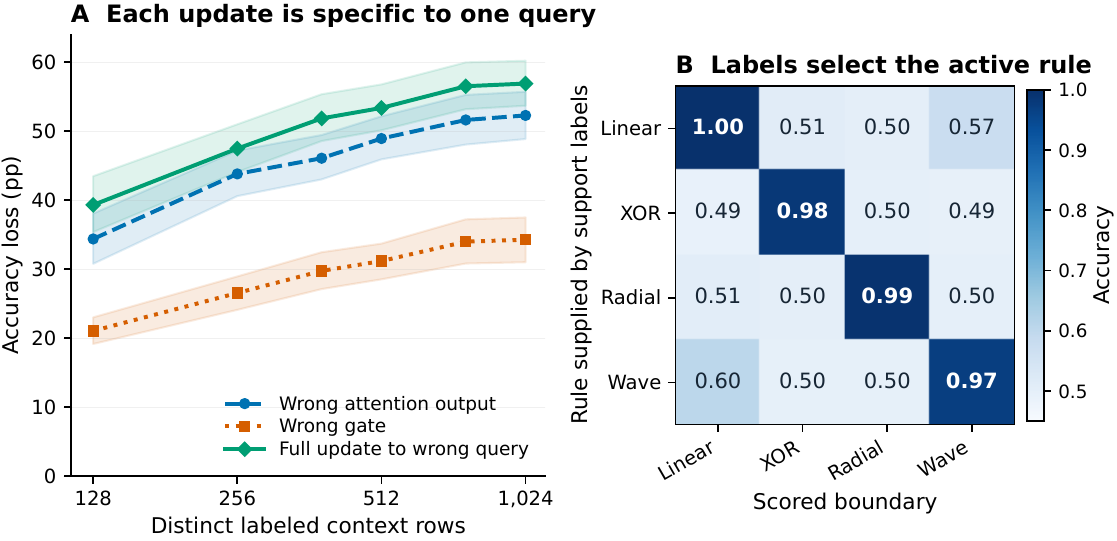}
\caption{\textbf{Controlled tests of context-dependent computation.}
\textbf{A:} Permuting attention, gate, or paired updates across queries tests
query specificity (paired 95\% CIs; 54 cells per context size).
\textbf{B:} Alternative support-label rules on the same points test rule selection.}
\label{fig:context-causality}
\label{fig:token-alignment-factorial}
\vspace{-3pt}
\end{figure}

The same points can require different boundaries. We now change only the
support labels to test this adaptation, then reassign internal updates to test
how the selected rule reaches each query (Figure~\ref{fig:context-causality}).

\begin{figure}[H]
\centering
\includegraphics[width=\textwidth]{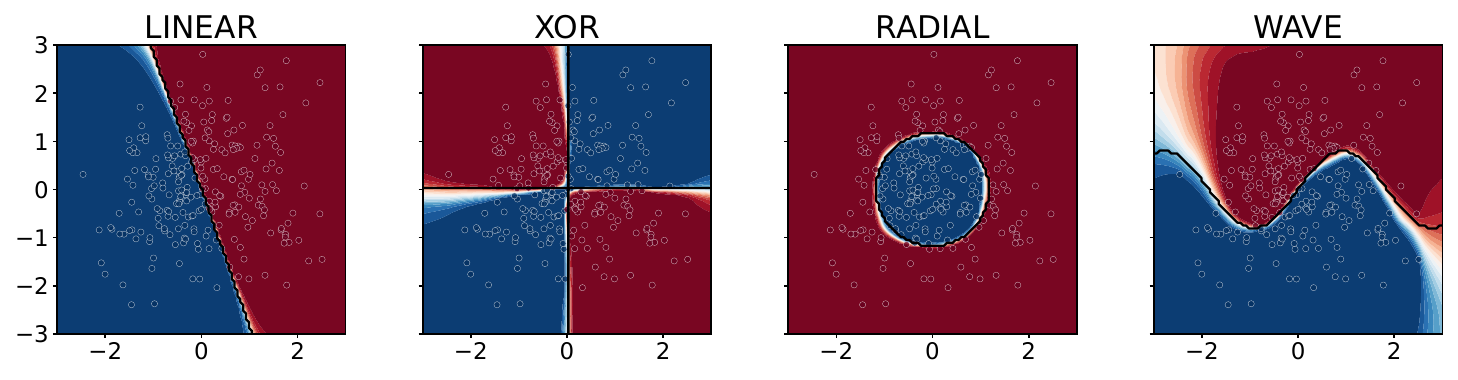}
\caption{\textbf{Support labels select the predictor in situ.} Features,
support coordinates, query coordinates, and model weights are identical across
panels. Only the support-label rule changes. The colored surfaces are
\modelname-L24 class-1 predictions; black curves show the generating boundaries.
The corresponding rule-selection test is Figure~\ref{fig:context-causality}B.}
\label{fig:context-programming}
\end{figure}

\paragraph{Labels select which predictor the model uses (Figure~\ref{fig:context-causality}B).}
On identical support and query coordinates, we change only support labels among
linear, XOR, radial, and wave rules, then score against all four query labelings.
Matching cells average 0.985 accuracy versus 0.515 off the diagonal:
labels change the predicted boundary on the same points
(Figure~\ref{fig:context-programming}).

\paragraph{Each correction is specific to its query (Figure~\ref{fig:context-causality}A).}
We permute the attention output, gate, or their paired update across queries
before residual addition. The paired test preserves both vectors but assigns
them to the wrong query. At 1,024 rows, accuracy drops by 52.28, 34.27, and
56.89 points, respectively. Even an intact update fails when sent to the wrong
query: its effect depends on the state receiving it.
Component recovery appears in Appendix~\ref{sec:component-recovery-protocol}.

\paragraph{Conclusion.}
Learning from a table includes improving the representations of the examples
themselves. Our derivation shows how labels define a support correction and
similarity transfers it to queries; interventions show that later predictions
benefit from evolving support states.
This motivates \modelname{}'s attention-gated, FFN-free stack, which reaches
1644.8 TabArena Elo and improves all four reported metrics over TabPFN-v3 on
both TabZilla views. These findings support repeated refinement of the current
task's representations, with attention and gating supplying contextual
nonlinearity without a separate expanded FFN.

\vspace{0.8em}
\subsection*{AI use statement}
Generative AI tools assisted with the conceptual framing of the work,
presentation of the derivation, experimental design and result interpretation,
software development, figure preparation, and manuscript editing. The authors
take responsibility for the mathematical claims, experimental results,
citations, code, and final manuscript, including AI-assisted content.

\clearpage
\begingroup
\sloppy
\setlength{\emergencystretch}{3em}
\bibliography{references}
\bibliographystyle{iclr2027_conference}
\endgroup

\clearpage
\appendix
\section*{Appendix: questions and evidence}
\addcontentsline{toc}{section}{Appendix: questions and evidence}
The main text follows one question: how can a fixed network construct a
predictor from a new labeled table? The appendix develops each part of the
answer, from the support correction and its query extension to the network,
benchmark comparisons, and controlled tests. The final experiments examine
when refinement is limited by the available evidence or by the readout used
to measure it.
\begin{center}
\small
\begin{tabularx}{\linewidth}{@{}Yp{0.24\linewidth}Y@{}}
\toprule
Question & Where to look & What the evidence establishes \\
\midrule
Does the derived update decrease the objective and extend to queries? &
Sections~\ref{sec:support-correction-details}--\ref{sec:functional-extension} &
Local improvement of the support objective and a controlled kernel approximation \\
Which operations are learned choices? & Section~\ref{app:implementation} &
Connection to attention, gating, and residual updates; full architecture \\
Do the benchmark comparisons use compatible settings? & Section~\ref{app:benchmarks} &
Dataset list, split and ensemble budgets; completed results \\
What do depth, working width and evidence change? & Section~\ref{sec:mechanism-protocols} &
Paired class-separation, dimension-limit, and context-size measurements \\
Are corrections episode- and query-specific? & Section~\ref{app:binding} &
Label switching, query permutation, and component recovery \\
Where does the refinement account fail? & Section~\ref{sec:supporting-mechanisms} &
Sparse identification, nonmonotone readouts and misleading raw rank \\
\bottomrule
\end{tabularx}
\end{center}

\section{Why the Support Update Decreases the Objective}
\label{sec:support-correction-details}

The question is whether the proposed support correction improves the stated
objective. We hold the current gradient and kernel fixed and assume local
smoothness along the update segment; the bound below concerns support geometry,
not unseen-label risk. For the exponential-dot-product kernel,
the columns of $F=-\nabla_Z\mathcal{J}(Z)$ are
\begin{equation}
 f_i=\frac{\tau}{n}\sum_{j\ne i}
 \left[
 a_{ij}\!\left(\frac{\mathbb{1}[y_i=y_j]}{q_i}-1\right)
 +a_{ji}\!\left(\frac{\mathbb{1}[y_i=y_j]}{q_j}-1\right)
 \right]z_j.
 \label{eq:support-gradient-expanded}
\end{equation}
Both outgoing and incoming neighborhood weights contribute because changing
$z_i$ changes its own prediction and those of the other support rows.

Let $U\in\mathbb{R}^{d\times n}$ be a candidate support-state correction,
$G=\nabla_Z\mathcal{J}(Z)=-F$, and
$K=[\kappa(z_i,z_j)]_{ij}+\epsilon I_n\succ0$ for $\epsilon>0$.
With $\alpha>0$, the local objective uses the Frobenius inner product
$\langle A,B\rangle=\operatorname{tr}(A^\top B)$ to trade first-order loss
change against update size and variation over the support geometry:
\begin{equation}
 \mathcal{Q}(U)=\underbrace{\langle G,U\rangle}_{\text{first-order loss change}}
 +\underbrace{\frac{\beta}{2}\lVert U\rVert_F^2}_{\text{local step size}}
 +\underbrace{\frac{\alpha}{2}\operatorname{tr}(UK^{-1}U^\top)}_{
 \text{smooth contextual correction}}.
 \label{eq:surrogate-objective}
\end{equation}
Here $\beta>0$ is the loss smoothness constant and $I_n$ is the $n\times n$
identity.
The kernel and the gradient are fixed at the current representation for this
local problem. Its first-order condition is
$F=\beta U^*+\alpha U^*K^{-1}$, giving
$U^*=FK(\beta K+\alpha I_n)^{-1}$. Moreover,
\begin{equation}
 \langle F,U^*\rangle
 =\beta\lVert U^*\rVert_F^2
 +\alpha\operatorname{tr}(U^*K^{-1}U^{*\top})\ge0.
\end{equation}
Smoothness along the update segment gives
\begin{align}
 \mathcal{J}(Z+U^*)-\mathcal{J}(Z)
 &\le-\langle F,U^*\rangle+\frac{\beta}{2}\lVert U^*\rVert_F^2\\
 &\le-\tfrac12\langle F,U^*\rangle,
\end{align}
which proves Equation~\ref{eq:descent-bound}: the derived support correction
decreases the episode objective locally.

\section{Extending Support Updates to Unseen Queries}
\label{sec:functional-extension}

Equation~\ref{eq:query-update-exact} follows from a vector-valued RKHS problem,
which makes explicit why a correction learned on support representations can be
evaluated at an unseen query. Let
$\mathcal{H}_{\widetilde\kappa}^{d}$ be the product RKHS on indexed states
$\bar z=(z,t)$, with
$\widetilde\kappa((z,t),(z',t'))=\kappa(z,z')+\epsilon\mathbb{1}[t=t']$.
Each support row and unseen query has a distinct index. Thus the support Gram
matrix is exactly $K$ with its diagonal term, while unseen-query cross-kernels
contain no such term. This convention also covers repeated feature vectors. Let
$g:(\mathbb{R}^{d}\times\mathcal I)\rightarrow\mathbb{R}^{d}$, writing $g(z_i)$
for its evaluation at support index $i$. For the support corrections
$F=[f_1,\ldots,f_n]$, consider
\begin{equation}
 \min_{g\in\mathcal{H}_{\widetilde\kappa}^{d}}
 -\sum_{i=1}^{n}\langle f_i,g(z_i)\rangle
 +\frac{\beta}{2}\sum_{i=1}^{n}\lVert g(z_i)\rVert_2^2
 +\frac{\alpha}{2}\lVert g\rVert_{\mathcal{H}_{\widetilde\kappa}^{d}}^2.
 \label{eq:functional-update-objective}
\end{equation}
By the representer theorem, the minimizer has the form
$g(\cdot)=\Gamma k_{\cdot}$, where
$k_z=[\widetilde\kappa(\bar z,\bar z_1),\ldots,\widetilde\kappa(\bar z,\bar z_n)]^\top$ and
$\Gamma\in\mathbb{R}^{d\times n}$. Substitution gives
\begin{equation}
 -\operatorname{tr}(F^\top\Gamma K)
 +\frac{\beta}{2}\lVert\Gamma K\rVert_F^2
 +\frac{\alpha}{2}\operatorname{tr}(\Gamma K\Gamma^\top).
\end{equation}
Differentiating and using $K\succ0$ yields
\begin{equation}
 \Gamma=F(\beta K+\alpha I_n)^{-1}
 =\frac{1}{\alpha}F\left(I_n+\frac{\beta}{\alpha}K\right)^{-1}.
\end{equation}
The indexed kernel gives the same update rule on support points and unseen queries.
Consequently $g(z)=\Gamma k_z$ is exactly
Equation~\ref{eq:query-update-exact}, and on support points
$[g(z_1),\ldots,g(z_n)]=\Gamma K=U^*$. Thus the query formula agrees with the
derived support update on the support points.

\paragraph{When is the attention-form approximation accurate?}
The resolvent identity and $K\succeq0$ give
\begin{equation}
 \left\|u(z_q)-\alpha^{-1}Fk_q\right\|_2
 \le \frac{\beta}{\alpha^2}\|F\|_2\|K\|_2\|k_q\|_2.
\end{equation}
Thus $\beta\|K\|_2/\alpha\ll1$ controls the absolute remainder relative to the
scale $\|F\|_2\|k_q\|_2/\alpha$. This bound motivates the normalized
readout and state-dependent scaling in Equation~\ref{eq:context-update}.

\section{From the Analytic Update to the Implemented Network}
\label{app:implementation}

We next ask how a network can learn the derived form across tables.
Table~\ref{tab:derivation-architecture} separates the required computations from
our implementation choices. Attention uses learned similarities
and contextual values; its normalized output is multiplied by a SiLU gate and
projected into a residual correction. This realizes evidence selection,
state-dependent scaling, and residual refinement in one block. The gate allows
signed, featurewise corrections rather than reconstructing the positive scalar
kernel mass $s(z_q)/\alpha$ in Equation~\ref{eq:query-update-factorized}.

\begin{figure*}[!htbp]
\centering
\includegraphics[width=\textwidth]{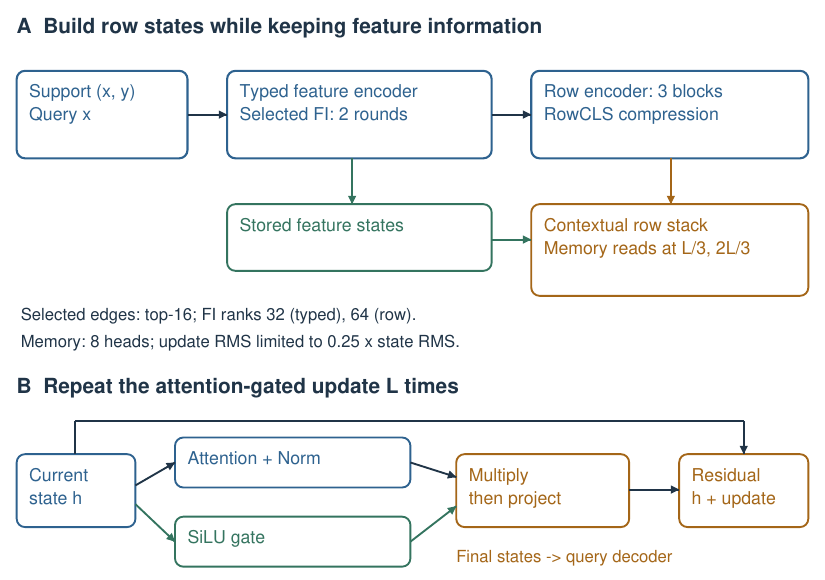}
\caption{\textbf{Feature and context flow through \modelname{}.}
\textbf{A:} RowCLS compresses the feature states while typed memory bypasses
compression and re-enters the contextual stack twice. \textbf{B:} Attention and
gating share the current state; their projected product is added to the residual.
The diagram shows the computation, with projection and normalization details in
Equation~\ref{eq:context-update}.}
\label{fig:refineicl-implementation-topology}
\end{figure*}

Figure~\ref{fig:refineicl-implementation-topology} shows two additional routes for
feature information. Selected feature interaction acts before compression;
typed memory later retrieves stored feature states. These components support the
architecture in Section~\ref{sec:refineicl-method}, while their empirical necessity
is assessed by deletion and recovery in Section~\ref{sec:component-recovery-protocol}.

\begin{table}[!htbp]
\caption{\textbf{From the derived update to \modelname{}.} The equations motivate roles; they do not equate learned values or gates with analytic gradients or kernel mass.}
\label{tab:derivation-architecture}
\centering
\footnotesize
\setlength{\tabcolsep}{3pt}
\renewcommand{\arraystretch}{0.96}
\begin{tabularx}{\textwidth}{@{}>{\raggedright\arraybackslash}X>{\raggedright\arraybackslash}X>{\centering\arraybackslash}p{0.13\textwidth}@{}}
\toprule
Derived term or operation & Network implementation & Source \\
\midrule
Similarity $\kappa(z,z_i)$ selects support rows & Attention weights $A_\ell$
& Eq.~\eqref{eq:query-update-attention} \\
Support labels and current error define $f_i$ & Learned attention values; not explicit gradients
& Eqs.~\eqref{eq:loo-objective}, \eqref{eq:query-update-attention} \\
Scale the update for the current query state
& SiLU gate & Eq.~\eqref{eq:context-update} \\
Apply $u(z)$ to the current representation & Projected residual update
& Eqs.~\eqref{eq:query-update-exact}, \eqref{eq:context-update} \\
Repeat successive updates & Repeated contextual blocks
& Eq.~\eqref{eq:context-update} iterated \\
The update depends nonlinearly on evidence and state & Expanded FFN omitted; tested empirically
& Eqs.~\eqref{eq:query-update-attention}--\eqref{eq:context-update} \\
\bottomrule
\end{tabularx}
\end{table}

\FloatBarrier

\section{Benchmark Protocols and Complete Comparisons}
\label{app:benchmarks}

\FloatBarrier
\subsection{TabArena v0.1 classification snapshot}

The exact 38 datasets are: APSFailure, Amazon employee access, Bank Customer Churn, Bioresponse, Diabetes130US, E-Commerce Shipping Data, Fitness Club, Give Me Some Credit, HR Analytics Job Change, Is-this-a-good-customer, MIC, Marketing Campaign, NATICUSdroid, SDSS17, anneal, bank-marketing, blood-transfusion-service-center, churn, coil2000 insurance policies, credit-g, credit card clients default, customer satisfaction in airline, diabetes, hazelnut-spread contaminant detection, heloc, hiva agnostic, in-vehicle coupon recommendation, jm1, kddcup09 appetency, maternal health risk, online shoppers intention, polish companies bankruptcy, qsar-biodeg, seismic-bumps, splice, students dropout and academic success, taiwanese bankruptcy prediction, and website phishing.

\modelname{} uses fold 0, eight ensemble members, and one support ordering shared
across model sizes. All 38 datasets are complete. This fixed benchmark snapshot contains 77
reference methods and is not interchangeable with the live leaderboard. The L24
row uses a binary-native continuation at step 20,000, distinct from the L24
checkpoint used for AMLB29. TabArena diagnostics informed this continuation's
prior choice; this is a development-exposed comparison. The separate
pure-Stage-2 checkpoint supplies the AMLB29 and extended OpenML results.

\FloatBarrier
\subsection{OpenML registries and splits}

We save the task data and repeat-0 split assignments before model evaluation.
OpenML-CC18 is suite 99 (72 tasks). AMLB29 contains the exact task IDs used by
TabPFNv2. The current Grinsztajn classification set combines categorical suite
334 (7 tasks) and numerical suite 337 (16 tasks). Historical22 retains all seven
categorical tasks and the fifteen numerical tasks reported by TabPFNv2; Clean18
removes OpenML task 361070 (the numerical \texttt{eye\_movements} leakage case),
its categorical counterpart, and duplicate datasets. Each result records all
views to which it belongs, so overlapping summaries are computed from the same
predictions rather than counted as new observations. Canonical TabZilla is suite
379 with 36 tasks. TabZilla102 is a separate, deterministic reconstruction of the
official 176-task set using the TabPFNv2 limits and normalized-name deduplication.
All ten official folds are saved for every task.

The views overlap and contain 183 unique task IDs in union. We average folds within
each task and then tasks within each named benchmark; overlapping memberships are
never treated as independent observations.

\FloatBarrier
\subsection{Native and common-budget inference}

The AMLB29 native protocol follows the archived TabPFNv2 classifier defaults: four estimators, seed 0, temperature 0.9, and a 10,000-row ceiling. We reproduce its multiclass AUC handling. If a test fold omits classes, the corresponding probability columns are removed, the remainder is renormalized, and targets are remapped before OVR-AUC is computed; a two-class remainder uses positive-class ROC-AUC. The extended common-budget panel also uses four estimators: OpenML-CC18 uses seed 42, identical capped support indices, and all 720 folds, while Grinsztajn and TabZilla use seed 0 and their complete registered folds. Query batching may change only the number of queries processed at once; support rows, preprocessing state, feature masks, and ensemble members remain unchanged. Differences in dataset lists and splits motivate separate tables rather than a pooled score.

The AMLB29 TabPFN-v2, PHE, and AutoGluon comparator rows are parsed from the
Nature supplement rather than recomputed through an unpinned package
constructor. We therefore label the baseline ``TabPFN-v2 (Nature default)''
throughout; it is not a claim about whichever model a current TabPFN release
selects by default.

\begingroup\raggedright
The current-V3 AMLB29 row is recorded as a direct evaluation with \texttt{tabpfn==8.0.7}
and checkpoint \texttt{tabpfn-v3-classifier-v3\_default.ckpt} (SHA-256
\texttt{d0d865d54dfbc524f5703104be90620182dca7e5fb2c16de72e9959ea18f3988}).
The checkpoint is loaded from a fixed local path. To isolate model
quality from ensemble budget, we use the same four estimators, seed 0,
temperature 0.9, 10,000-row support ceiling, folds, and externally materialized
data as the other AMLB29 rows. This is the benchmark's standard four-member
comparison, not a claim about the package's native estimator default. All 290
folds are recorded as complete without failure.
\par\endgroup

A separate TabICLv2 CC18 reference uses release \texttt{v2.0.0}, eight members,
seed 42, and a 10,000-row support ceiling, with 720/720 completed folds. It is
reported separately from the four-member panel. A runtime
compatibility patch changes two equivalent tuple reductions but not model outputs.

\FloatBarrier
\subsection{Available evaluation records}
\label{app:benchmark-provenance}

The AMLB29 L12/L24 summaries report 290 complete folds each. Reaggregation of
the 290 TabPFN-v3 fold records gives 0.868765 accuracy and 0.936141 OVR-AUC.
The L24 AMLB29 checkpoint SHA-256 is
\texttt{071de368d16d137595d59249a7837d32}\allowbreak
\texttt{4daf97733acc1d1c5c9f90fd3c5db22b}.
TabArena uses a later L24 checkpoint (SHA-256
\texttt{1a96b0ce5f6b1e99a24eb50b65ec20d2}\allowbreak
\texttt{5af042fd1cae08b196ea7703a5a2e12f}).
Its 38 per-task outputs and the same 77 reference methods reproduce 1644.8
Elo, 11.078947 mean rank, 0.869105 win rate, and 0.162385 mean error.
The six CC18, Grinsztajn, and TabZilla views in
Table~\ref{tab:benchmark-breadth-ffn}A were also reaggregated from complete
ten-fold results for L12, L24, and TabPFN-v3. All records in each view use a
single protocol ID and checkpoint hash. The eight-member TabICLv2 CC18
reference is reported separately from these four-member comparisons.
The supplement includes archived fold metrics and a script that reconstructs
the paired dataset averages in Section~\ref{app:paired-benchmarks}, checking
task/fold identities, checkpoint hashes, and coverage before aggregation.

\subsection{Aggregate benchmark comparisons}
\label{app:extended-public-tables}

The tables below compare \modelname-L12 and \modelname-L24 with the current
TabPFN-v3 checkpoint under the matched four-member protocols defined in
Appendix~\ref{app:benchmarks}. Each number is a dataset-macro mean over the
complete registered split set; overlapping benchmark views reuse the same
predictions and are not pooled into a single score. The CC18, Grinsztajn, and
TabZilla means were recomputed from complete fold results
(Section~\ref{app:benchmark-provenance}).

\begin{table}[!htbp]
\caption{Public benchmark coverage and inference settings. Overlapping OpenML views are intentional and are aggregated separately.}
\label{tab:protocols}
\centering
\small
\setlength{\tabcolsep}{3.5pt}
\begin{tabularx}{\linewidth}{lrrYY}
\toprule
Benchmark view & Data & Splits & Inference setting & Primary statistics \\
\midrule
TabArena v0.1 snapshot & 38 & fold 0 & 8 members & Elo/rank/win rate \\
AMLB classification & 29 & all folds & native 4, 10K rows & OVR-AUC/accuracy \\
OpenML-CC18 & 72 & all folds & common 4 & OVR-AUC/accuracy \\
Grinsztajn current / historical & 23 / 22 & all folds & native 4 & accuracy/AUC \\
TabZilla canonical / extension & 36 / 102 & all folds & native 4 & loss/AUC/accuracy/F1 \\
\bottomrule
\end{tabularx}
\end{table}
\FloatBarrier

\begin{table}[!htbp]
\caption{OpenML-CC18 with the same four-member inference budget. Every reported model receives identical capped support rows and uses neither HPO nor dataset-specific gradient updates. We average folds within each dataset and then datasets.}
\label{tab:openml-public}
\centering
\small
\setlength{\tabcolsep}{3.0pt}
\begin{tabular*}{\linewidth}{@{\extracolsep{\fill}}lrrr@{}}
\toprule
Model & Accuracy $\uparrow$ & OVR-AUC $\uparrow$ & Log loss $\downarrow$ \\
\midrule
\modelname-L12 & 0.86629 & 0.93278 & 0.33101 \\
\modelname-L24 & 0.88187 & \textbf{0.93832} & 0.28305 \\
TabPFN-v3 (current checkpoint) & \textbf{0.88600} & 0.93704 & \textbf{0.27409} \\
\bottomrule
\end{tabular*}
\end{table}
\FloatBarrier

The four-member CC18 comparison favors TabPFN-v3 for accuracy and log loss and
L24 for OVR-AUC. A separately completed \emph{eight-member} TabICLv2 v2.0.0
reference reaches 0.88930 accuracy, 0.93864 OVR-AUC and 0.25615 log loss on the
same 72-dataset list (720 folds, zero unresolved failures). Its different
member budget precludes inclusion in the four-member winner marking.

\begin{table}[!htbp]
\caption{Grinsztajn classification under four-member native inference. Current23 is reported jointly and by feature type; Historical22 and leakage-safe Clean18 are sensitivity views. We average folds within each dataset and then datasets.}
\label{tab:grinsztajn-public}
\centering
\small
\setlength{\tabcolsep}{3.5pt}
\begin{tabular*}{\linewidth}{@{\extracolsep{\fill}}lll@{\hspace{8pt}}rr@{}}
\toprule
View & Group & Model & Accuracy $\uparrow$ & ROC-AUC $\uparrow$ \\
\midrule
\multirow{9}{*}{Current23} & All (23) & \modelname-L12 & 0.78106 & 0.84928 \\
 & & \modelname-L24 & \textbf{0.81762} & \textbf{0.88341} \\
 & & TabPFN-v3 & 0.79917 & 0.86635 \\
\cmidrule(lr){2-5}
 & Categorical (7) & \modelname-L12 & 0.73544 & 0.80594 \\
 & & \modelname-L24 & \textbf{0.79665} & \textbf{0.86365} \\
 & & TabPFN-v3 & 0.76436 & 0.83633 \\
\cmidrule(lr){2-5}
 & Numerical (16) & \modelname-L12 & 0.80102 & 0.86823 \\
 & & \modelname-L24 & \textbf{0.82679} & \textbf{0.89206} \\
 & & TabPFN-v3 & 0.81440 & 0.87948 \\
\midrule
\multirow{6}{*}{Historical22} & All (22) & \modelname-L12 & 0.78730 & 0.85579 \\
 & & \modelname-L24 & \textbf{0.81522} & \textbf{0.88062} \\
 & & TabPFN-v3 & 0.80424 & 0.87140 \\
\cmidrule(lr){2-5}
 & Numerical (15) & \modelname-L12 & 0.81151 & 0.87905 \\
 & & \modelname-L24 & \textbf{0.82389} & \textbf{0.88854} \\
 & & TabPFN-v3 & 0.82285 & 0.88776 \\
\midrule
\multirow{3}{*}{Clean18} & All (18) & \modelname-L12 & 0.79391 & 0.86116 \\
 & & \modelname-L24 & \textbf{0.80751} & \textbf{0.87232} \\
 & & TabPFN-v3 & 0.80669 & 0.87176 \\
\bottomrule
\end{tabular*}
\end{table}
\FloatBarrier

The Grinsztajn sensitivity views ask whether the observed L24 advantage survives
historical membership and leakage/duplicate exclusions. Read each view separately:
L24 remains ahead in both metrics, but its Clean18 advantage is only 0.00082
accuracy and 0.00056 AUC. The unadjusted mean differences support consistency of
the sign, not a claim of statistical separation.

\begin{table}[!htbp]
\caption{TabZilla under four-member native inference. Canonical36 is OpenML suite 379. Extension102 reproduces the separate TabPFNv2 filtering and deduplication view.}
\label{tab:tabzilla-public}
\centering
\small
\setlength{\tabcolsep}{3.5pt}
\begin{tabular*}{\linewidth}{@{\extracolsep{\fill}}llrrrr@{}}
\toprule
View & Model & Accuracy $\uparrow$ & Macro F1 $\uparrow$ & AUC $\uparrow$ & Log loss $\downarrow$ \\
\midrule
\multirow{3}{*}{Canonical36} & \modelname-L12 & 0.85198 & 0.80301 & 0.91984 & 0.35953 \\
 & \modelname-L24 & \textbf{0.86760} & \textbf{0.82662} & \textbf{0.92820} & \textbf{0.31356} \\
 & TabPFN-v3 & 0.86152 & 0.81703 & 0.92466 & 0.33959 \\
\midrule
\multirow{3}{*}{Extension102} & \modelname-L12 & 0.87633 & 0.82707 & 0.91893 & 0.27583 \\
 & \modelname-L24 & \textbf{0.88867} & \textbf{0.84818} & \textbf{0.92773} & \textbf{0.25045} \\
 & TabPFN-v3 & 0.88525 & 0.84106 & 0.92268 & 0.26339 \\
\bottomrule
\end{tabular*}
\end{table}
\FloatBarrier

Canonical36 uses the official TabZilla dataset list; Extension102 checks sensitivity
to the separate filtering/deduplication reconstruction. L24 improves all four
reported metrics against TabPFN-v3 in both views. The overlapping memberships
are not independent replications.

\begin{table}[!htbp]
\caption{Model capacity reference; parameter counts do not equal matched inference cost. BF16 storage counts raw model
weights; AutoGluon is an ensemble rather than a single checkpoint.}
\label{tab:capacity-reference}
\centering
\small
\setlength{\tabcolsep}{4.8pt}
\begin{tabular}{lrrrr}
\toprule
Model & Blocks & Width & Parameters & BF16 weights \\
\midrule
\modelname-L12 & 12 & 2048 & 317.9M & 0.592 GiB \\
\modelname-L24 & 24 & 1024 & 159.5M & 0.30 GiB \\
TabFM v1.0.0 & 24 & 2048 & 1.639B & 3.05 GiB \\
TabPFN-v3 & 24 & 128 & 53.2M & 0.099 GiB \\
TabICLv2 & 12 & 128 & 27.6M & 0.051 GiB \\
TabPFN-v2.6 & 12 & 192 & 7.2M & 0.013 GiB \\
AutoGluon 1.5 & -- & -- & Ensemble & -- \\
\bottomrule
\end{tabular}
\end{table}
\FloatBarrier

\FloatBarrier

\subsection{How broadly are the gains distributed across datasets?}
\label{app:paired-benchmarks}

Benchmark means can hide differences in which datasets favor each model.
We therefore pair L24 and TabPFN-v3 on the same ten repeat-0 folds, average
within each dataset, and compare the resulting dataset scores. All paired
dataset and fold identifiers agree. Table~\ref{tab:paired-benchmarks} reports
mean differences and wins/ties/losses; a tie is an absolute difference at most
$10^{-12}$. We resample whole datasets 20,000 times, retaining the pairing
between models and all folds. The 95\% percentile intervals describe variation
across datasets in each view. They are pointwise intervals, not adjusted for
multiple comparisons; overlapping views are analyzed separately, never pooled.

\begin{table}[!htbp]
\centering
\small
\setlength{\tabcolsep}{3pt}
\caption{\textbf{Dataset-level differences against TabPFN-v3.} Differences
are percentage points (L24 minus v3), with paired 95\% intervals. W/T/L counts
datasets, not folds. Both models use four members.}
\label{tab:paired-benchmarks}
\begin{tabular}{lrrrr}
\toprule
View & Accuracy difference & W/T/L & AUC difference & W/T/L \\
\midrule
OpenML-CC18 & $-0.41$ [$-1.25$, $+0.28$] & 44/7/21 & $+0.13$ [$-0.16$, $+0.34$] & 51/5/16 \\
Grinsztajn Current & $+1.84$ [$+0.04$, $+4.50$] & 13/0/10 & $+1.71$ [$+0.04$, $+4.14$] & 16/0/7 \\
Grinsztajn Clean18 & $+0.08$ [$-0.002$, $+0.17$] & 10/0/8 & $+0.06$ [$-0.009$, $+0.13$] & 12/0/6 \\
TabZilla Canonical & $+0.61$ [$-0.44$, $+1.60$] & 23/2/11 & $+0.35$ [$-0.25$, $+0.95$] & 27/2/7 \\
TabZilla Extension & $+0.34$ [$+0.09$, $+0.62$] & 52/19/31 & $+0.51$ [$+0.21$, $+0.93$] & 68/17/17 \\
\bottomrule
\end{tabular}

\end{table}

On TabZilla Extension, L24's mean gains are 0.34 accuracy points
[0.09, 0.62] and 0.51 AUC points [0.21, 0.93], with AUC improving on 68
datasets and declining on 17. Canonical AUC improves on 27 datasets and
declines on seven, although its mean-difference interval includes zero.
CC18 illustrates why the win count and mean answer different questions:
accuracy improves on 44 datasets but the mean favors v3, indicating larger
losses on some tasks. Grinsztajn Clean18 retains small positive means with
intervals spanning zero. Thus the broadest paired evidence here is the
TabZilla Extension result, rather than a count of overlapping benchmark means.

\section{How Depth, Width, Context, and the FFN Affect the Model}
\label{sec:mechanism-protocols}

The refinement view connects model capacity to a concrete computation:
updating task representations from labeled rows. We first examine whether
learned updates improve those representations and help later queries. We then
limit their dimensions, vary their evidence, and measure the prediction and
memory trade-off of adding an FFN. Together these tests ask what resources
the computation uses, and what additional expansion buys.

\FloatBarrier
\subsection{Do later blocks improve class separation for the current task?}
\label{app:depth-protocol}

\paragraph{Setup and measured quantity.}
Figure~\ref{fig:three-resources}A uses the pretrained L24 15K checkpoint without
parameter updates and a held-out
set of $3$ class counts ($2,4,8$), $3$ RBF component counts ($4,16,64$), and $8$
replicates: 72 episodes, each with 8 features, 1,024 support and 256 query rows.
A 256-landmark centered RBF kernel defines equal-energy Gaussian class scores in
its first $B$ nonconstant eigenmodes; query scores use a Nystr\"om extension.
The kernel bandwidth is the support-landmark median distance. The held-out seed
805843 changes coordinates, teacher coefficients and labels. Each class must
occupy at least 3\% of support rows. No probe is fitted to the test episodes.

At each block, normalize row states, average support states within each class,
and normalize the resulting prototypes. The two paired measures are the query
correct-minus-nearest-wrong cosine margin and nearest-prototype accuracy at
block 24 minus block 12. Bootstrap units are the 72 episodes, keeping both layers
paired. The margin gain is 0.0509 [0.0403, 0.0616]; the accuracy gain is 2.77
[1.40, 4.02] percentage points. Both support continued class-level refinement.
The net improvement from block 12 to block 24 shows stronger class geometry
after repeated contextual updates.

\FloatBarrier
\subsection{Do the learned block updates improve the support objective?}
\label{sec:learned-support-update}

Figure~\ref{fig:three-resources-v2}A measures class geometry, but does not
directly test the objective that motivates the derivation. We therefore measure
Equation~\ref{eq:loo-objective} immediately before and after each of the 24
contextual blocks of the same L24 checkpoint. This follow-up diagnostic reuses
the 72 registered geometry episodes (three class counts, three rule bandwidths, eight
replicates), each with 1,024 labeled support rows. We fix
$\tau=1/\sqrt{1024}$ for every layer and task. As a control, we permute a
block's learned support-row updates before adding them to its input states;
this retains the set of update vectors but breaks their row assignment. A
second control shuffles support labels when evaluating the objective. Neither
control changes model parameters or generates a new forward trajectory.

\begin{table}[H]
\centering
\small
\setlength{\tabcolsep}{6pt}
\caption{\textbf{Learned support updates and the leave-one-out objective.}
Entries are the change in Equation~\ref{eq:loo-objective}, averaged first
over 24 blocks and then over 72 episodes. Negative is better. Intervals
bootstrap episodes, not blocks; controls are one-block counterfactuals
evaluated at each actual block input. The final column counts episodes whose
mean block change is negative.}
\label{tab:learned-support-update}
\begin{tabular}{lrr}
\toprule
Update assignment & Mean $\Delta\mathcal{J}$ per block & Episodes $<0$ \\
\midrule
Actual learned update & $-0.0204$ [$-0.0226$, $-0.0183$] & 72/72 \\
Permuted support-row update & $+0.0258$ [$+0.0222$, $+0.0300$] & 0/72 \\
Shuffled labels for scoring & $+0.6312$ [$+0.5297$, $+0.7384$] & 0/72 \\
\bottomrule
\end{tabular}
\end{table}

The actual trajectory lowers this support objective, whereas both controls
raise it. This links the trained stack to the label-defined geometric target
more directly than prototype separation alone. To remove state magnitude as an
explanation, we also normalize each support state to unit length and score the
same leave-one-out objective with $\tau=1$: mean block changes are $-0.00458$
for actual updates and $+0.01192$ for row-permuted updates (72/72 episodes
favor the actual assignment). It is not a literal gradient
trajectory: the mean cosine between a block update and the negative objective
gradient is only $0.0071$, and some blocks raise the objective. The result
identifies a stack-level effect on this surrogate, not the exact operator in
Equation~\ref{eq:support-update}.

Does a learned support write matter to the query prediction? In a separate
paired intervention on the same episodes, we keep block 12's query output
unchanged but replace its support output with its input, then run the remaining
blocks normally. We score cross-entropy over the task's $C$ class logits.
Skipping that one support write raises final query CE by 0.0510 on average
(episode-bootstrap 95\% CI [0.0417, 0.0608]; 72/72 episodes) and lowers query
accuracy by 1.44 points. The intervention shows a causal contribution of the
mid-stack support write to query prediction. It does not establish that the
leave-one-out objective mediates that contribution: after centering within
each of the nine class-count/rule-bandwidth cells, the Spearman correlation
between block-12 objective reduction and query-CE benefit is $-0.026$.
The supplement includes the paired episode records, probe drivers, and policies.

\subsection{Do broader rules use more representation dimensions?}
\label{sec:width-dose-table}

\paragraph{Setup and measured quantity.}
Figure~\ref{fig:three-resources}B fixes the same L24 checkpoint, eight features,
1,024 support rows and 256 queries. The teacher construction above varies only
$B\in\{2,4,8,16,32,64,128\}$ across three class counts and six replicates
(seed 805742). A single rank-32 PCA basis, calibrated on disjoint episodes,
projects the 1,024-dimensional contextual state after block 4; all later blocks
remain unchanged. The primary comparison is the bottleneck loss at $B=64$ minus
that at $B=4$ within each of the 18 class-count/replicate units.
Full-state accuracy must exceed chance by 10 points in every primary
bandwidth-by-class cell; this validity check passes.

Table~\ref{tab:width-dose-full} reports all seven rule sizes. Read the increase in
full-minus-projected accuracy loss, not raw hidden rank. The primary increase is
7.62 [6.03, 9.05] points. The result shows task-dependent use of representation
dimensions within this checkpoint; retraining a smaller-width network could
compensate. The smallest and largest rule sizes extend the original comparison;
they are not additional main tests. Within the selected $B$ kernel modes, label
$r_{90}$ counts the leading modes accounting for 90\% of the projected,
centered one-hot label energy; its denominator excludes the remaining modes.

\begin{table}[!htbp]
\centering
\small
\caption{Effect of the rank-32 projection across seven rule sizes. Loss is full-state accuracy minus
accuracy after projecting the state to the same rank-32 subspace.}
\label{tab:width-dose-full}
\begin{tabular}{lrrrrrrr}
\toprule
$B$ & 2 & 4 & 8 & 16 & 32 & 64 & 128 \\
\midrule
Within-$B$ label $r_{90}$ & 1.9 & 3.8 & 7.3 & 14.2 & 28.8 & 57.7 & 112.6 \\
Full accuracy & .9698 & .9674 & .9683 & .8987 & .8804 & .8214 & .7804 \\
Rank-32 loss (pp) & 1.06 & 1.04 & 1.65 & 5.30 & 8.42 & 8.66 & 11.05 \\
\bottomrule
\end{tabular}
\end{table}

\FloatBarrier
\subsection{Does additional distinct evidence change the architecture gap?}
\label{app:evidence-protocol}

\paragraph{Setup and measured quantity.}
Figure~\ref{fig:three-resources}C uses L12 at 40K and L24 at 15K. The experiment has
three boundary families (checkerboard, rings, wave), six frequencies
($2,4,6,8,10,12$) and four replicates, for 72 paired conditions (seed 804815).
Coordinates lie in $[-3,3]^2$. Each condition fixes 512 queries in a boundary band
of width 0.2 and a shared set of 1,024 support rows; nested prefixes of
128, 192, 256, 384, 512, 768 and 1,024 rows contain no duplicates.

The primary measure is the change in the L24-minus-L12 accuracy gap between 128
and 1,024 rows, with bootstrap resampling of all 72 paired conditions. The
4.01 [2.26, 5.79]-point interaction supports an increasing overall architecture
advantage with evidence. It is not pointwise monotone: the gap decreases from
1.02 points at 192 rows to 0.75 at 256. The checkpoints differ in depth, width,
and training history, so this experiment isolates the effect of evidence on
their gap, not an isolated effect of depth or parameter count.

\FloatBarrier
\subsection{Is an expanded FFN needed at a fixed training budget?}
\label{sec:depth-ffn-ablation}

The question is whether omitting an expanded FFN prevents competitive prediction
when optimization opportunity is held fixed. Table~\ref{tab:depth-ffn-ablation} reports the complete six-cell $d=96$
depth-by-FFN experiment after the same 100K training updates. All cells use
RowCLS8, 512-row episodes (30--90\% labeled support), global batch 32, the official GraphSCM-v2 Stage-1
classification prior, contextual gating, feature interaction, typed memory,
and the same optimizer and validation protocol. The global FFN setting changes
the column, row, and contextual blocks together. They share one training seed
(42); these are single-seed results, not averages over independent training runs.
The reported result is held-out GraphSCM-v2 validation, separate from the public
benchmarks and the fixed-checkpoint boundary-rule evaluation below.

\begin{table}[!htbp]
\centering
\small
\caption{\textbf{Fixed-budget depth-by-FFN ablation.} All six cells are reported
at 100K updates and $d=96$. Bold marks the better value within each depth and
metric. This comparison describes finite-budget behavior, not a converged
scaling law.}
\label{tab:depth-ffn-ablation}
\begin{tabular}{llrr}
\toprule
Depth & Model variant & Validation CE $\downarrow$ & Accuracy (\%) $\uparrow$ \\
\midrule
L2 & Gated no-FFN & \textbf{0.454457} & 81.6553 \\
   & Gated + FFN-2$\times$ & 0.458385 & \textbf{82.1154} \\
\midrule
L4 & Gated no-FFN & \textbf{0.440428} & \textbf{82.0533} \\
   & Gated + FFN-2$\times$ & 0.469747 & 81.8505 \\
\midrule
L8 & Gated no-FFN & 0.450115 & 82.0674 \\
   & Gated + FFN-2$\times$ & \textbf{0.438677} & \textbf{82.3395} \\
\bottomrule
\end{tabular}
\end{table}

At this budget, L4 no-FFN is within 0.00175 CE of L8 FFN-2$\times$.
The FFN effect depends on depth: it worsens CE at L2 and L4 but improves it
at L8; at L2, CE and accuracy favor different variants. The L4 no-FFN model
nearly matches the L8 FFN model's CE with a smaller contextual stack.

\FloatBarrier
\subsection{How much accuracy does the FFN buy at each memory cost?}
\label{sec:ffn-context-memory-protocol}

We measure the prediction--memory trade-off of whole-model FFN expansion using
the paired $d=96$, eight-block models at 100K updates. The pair shares the
initialization seed, GraphSCM-v2 episode stream, optimizer, eight RowCLS tokens,
contextual gating, feature-interaction modules, typed memory, and 512-row training episodes
with a 30--90\% support fraction. The shared training seed is 42; optimization
uses Muon, peak learning rate $8\times10^{-4}$, 500 warmup steps and cosine
decay to $8\times10^{-5}$. A global factor of zero or two disables or enables
FFNs in all six column-attention subblocks, three row-encoder blocks, and eight
contextual blocks. This compares whole-model FFN expansion, not a contextual-only
intervention. Total parameters increase from 30,934,796 to 50,154,956 (+62.1\%):
222,336 are added to the column encoder, 111,168 to the row encoder, and
18,886,656 to the contextual predictor. The supplement includes an executable
parameter audit. Measured memory also includes runtime allocations.

We keep both checkpoints unchanged and evaluate three boundary-rule families, six
frequencies and three replicates (54 paired conditions, seed 806031), with
384 fixed boundary-band queries per condition, at seven
nested context sizes, $n\in\{128,192,256,384,512,768,1024\}$. Every added row
is a distinct labeled observation; no row is duplicated to equalize token count.
For each condition, the support and query coordinates, labels, and nesting order
are identical across models. Memory is measured on an NVIDIA A100-SXM4-80GB,
with FP32 model weights and FP16 autocast, one episode per forward pass, and
all 384 queries in one batch. We use the raw tensor forward path, no ensemble,
zero dropout, and no gradient recording or activation recomputation. Before
each measurement we reset the CUDA peak-memory counter; the reported maximum
allocated memory includes resident weights and temporary tensors, not reserved
allocator memory. Paired bootstrap intervals resample the 54 conditions.

\begin{figure}[htbp]
\centering
\includegraphics[width=0.88\textwidth]{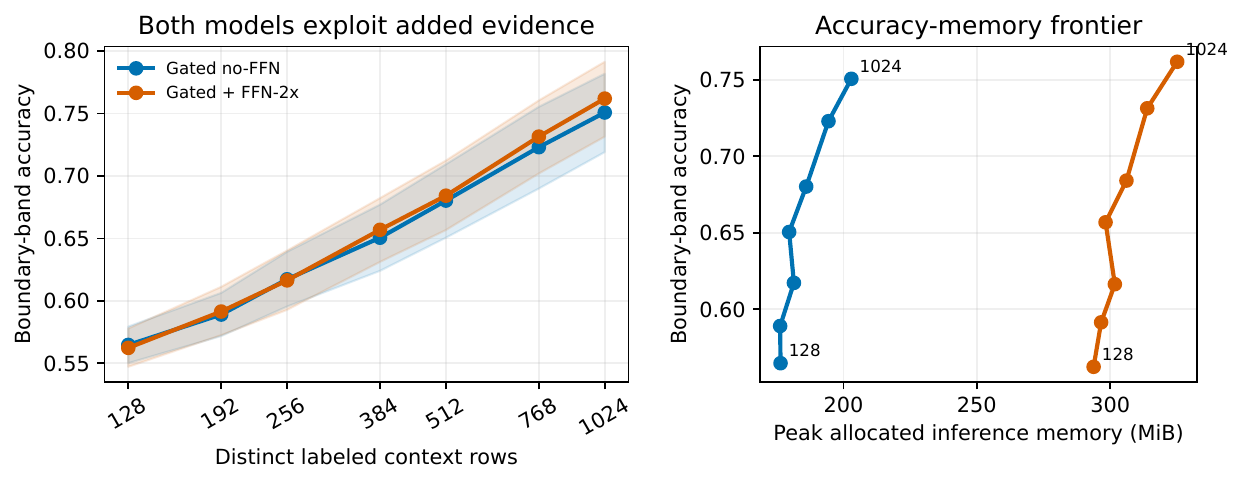}
\caption{\textbf{Accuracy and memory as labeled context grows, with and without an expanded
FFN.} Accuracy is measured on the same 54 boundary-rule conditions as the number
of distinct labeled rows increases. The right panel plots accuracy against peak allocated memory, including model
weights. A point is a context size, not a matched-memory retraining comparison.}
\label{fig:ffn-context-memory-frontier}
\end{figure}

Figure~\ref{fig:ffn-context-memory-frontier} shows the complete result underlying
Table~\ref{tab:benchmark-breadth-ffn}B. At 1,024 rows, the expanded FFN improves
accuracy by 1.11 points (95\% CI $[0.38,1.89]$), while increasing peak memory by
60.2\%. Both models improve strongly with distinct
evidence: 18.60 points without the expanded FFN and 19.95 points with it. The FFN also increases the 128--1,024 evidence gain by 1.35 points
[0.26, 2.45]. Thus it can improve evidence use, while the no-FFN model already
scales strongly. These paired-episode intervals condition on a single training
seed and do not measure variability across training runs.

\FloatBarrier

\section{How Support Labels and Queries Control the Update}
\label{app:binding}

Improved representations alone do not show how the model adapts to a particular
task. The following tests isolate two decisions: which rule the support labels
select, and which correction reaches each query. Component recovery then tests
how much the remaining network can compensate when part of this computation
is removed.

\FloatBarrier
\subsection{Do support labels select a different predictor?}

Figure~\ref{fig:context-programming} shows the predictions summarized by the
heatmap in Figure~\ref{fig:context-causality}B. With L24 parameters unchanged and identical
support/query coordinates, only support labels change among linear, XOR, radial
and wave rules. The output is the predicted class-1 probability surface; the
black generating boundary provides the reference. Matching-rule accuracy averages
0.98498 versus 0.51475 against other rules. Thus, on the same point cloud, the
support labels determine which predictor the model uses.

\FloatBarrier
\subsection{Does each query need its own attention output and gate?}
\label{sec:token-alignment-factorial}

The question is whether each correction must reach the query for which it was
computed (Figure~\ref{fig:context-causality}A). The experiment uses the pretrained L24 step-15K checkpoint without parameter updates and a separate paired RBF
episode set. Each context size contains 54 tasks spanning
three class counts, three RBF component counts, and six replicates; every condition
receives exactly the same support features, labels, queries, and targets. In the
evidence-mismatch condition, attention outputs are permuted across query tokens
while gates and residual states remain fixed. In the gate-mismatch condition,
gate preactivations are permuted while routed evidence and residual states remain
fixed. In the matched-update condition, evidence and gate preactivation receive
the same permutation, preserving their pairing while assigning the resulting
update to the wrong residual state. Every permutation preserves the set of vectors
within the query segment. Table~\ref{tab:token-alignment-full} reports accuracy
loss relative to the unchanged model, with paired 95\% bootstrap intervals.

\begin{table}[!htbp]
\centering
\small
\setlength{\tabcolsep}{4.2pt}
\caption{\textbf{Complete query-permutation experiment.} Accuracy damage in percentage
points after assigning one part of the contextual update to another query.}
\label{tab:token-alignment-full}
\begin{tabular}{rrrr}
\toprule
Context rows & Wrong attention output & Wrong gate & Full update $\rightarrow$ wrong query \\
\midrule
128   & 34.38 [30.78, 38.06] & 21.07 [19.17, 23.03] & 39.31 [35.38, 43.48] \\
256   & 43.81 [40.59, 47.19] & 26.56 [24.11, 28.95] & 47.47 [44.08, 50.99] \\
384   & 46.08 [43.00, 49.44] & 29.75 [27.12, 32.46] & 51.85 [48.57, 55.35] \\
512   & 48.93 [45.91, 52.15] & 31.18 [28.53, 33.71] & 53.36 [50.12, 56.77] \\
768   & 51.62 [48.05, 55.25] & 33.99 [30.82, 37.23] & 56.51 [53.18, 59.96] \\
1,024 & 52.28 [48.86, 55.70] & 34.27 [31.01, 37.49] & 56.89 [53.68, 60.20] \\
\bottomrule
\end{tabular}
\end{table}

\FloatBarrier

All three damage intervals remain positive at every context size. Even keeping
the evidence--gate pair intact does not rescue the wrong-recipient update,
showing that updates are query-specific.

\FloatBarrier
\subsection{Can the remaining components compensate after deletion?}
\label{sec:component-recovery-protocol}

Which components does the trained model depend on, and which recover after brief
adaptation? All conditions start from L24 at 15K. Direct deletion is evaluated at
step 0; recovery keeps the deletion, loads weights only, resets Muon optimizer
state, and uses the same GraphSCM-v2 stream for 250 updates (learning rate
$10^{-4}$, global batch 8, seed 918273). The unmodified reference receives the
same continuation. We report all three planned checkpoints (25, 100, and 250 updates).

The typed-frontend tests set one or both pre-compression residuals to zero; the
row-FI tests set the row-encoder adapters to zero and keep them fixed. The memory
tests remove one or both reads. The gate test replaces all 24 SiLU gate outputs with one while
preserving attention, normalization, projection and residual writes. These
operations measure component dependence in a trained model, not from-scratch
architectural necessity.

Table~\ref{tab:component-recovery-full} gives the full RBF results. Most individual
frontend and row-FI changes stay near zero, whereas removing both memory reads
continues to hurt and the gate effect remains large after recovery. At 250 updates,
removing the gate, memory, and row FI together costs 42.54 points, compared with
28.94 for the gate alone. This identifies an additional effect of jointly
removing memory and row FI; their individual effects are much smaller.

\begin{table}[!htbp]
\centering
\small
\setlength{\tabcolsep}{5.0pt}
\caption{Full matched-recovery results on the same RBF episode set. Entries are accuracy
damage in percentage points relative to the unmodified model after the same
number of training updates; negative values favor the modified model.}
\label{tab:component-recovery-full}
\begin{tabular}{lrrrr}
\toprule
Disabled component & Step 0 & Step 25 & Step 100 & Step 250 \\
\midrule
Typed frontend, first refinement & $-0.02$ & $-0.12$ & $-0.04$ & $0.12$ \\
Typed frontend, second refinement & $-0.09$ & $-0.45$ & $-0.04$ & $-0.21$ \\
Typed frontend, both refinements & $-0.15$ & $-0.47$ & $-0.05$ & $-0.19$ \\
Row FI, first adapter & $0.00$ & $-0.16$ & $0.13$ & $0.02$ \\
Row FI, second adapter & $0.06$ & $-0.12$ & $0.08$ & $-0.03$ \\
Row FI, both adapters & $0.18$ & $-0.01$ & $0.35$ & $0.23$ \\
Typed memory, first read & $0.74$ & $0.36$ & $0.87$ & $0.97$ \\
Typed memory, second read & $0.06$ & $-0.10$ & $0.37$ & $0.21$ \\
Typed memory, both reads & $0.94$ & $0.87$ & $1.10$ & $0.86$ \\
Typed memory + row FI & $1.52$ & $1.03$ & $1.35$ & $1.08$ \\
Contextual gate & $45.19$ & $42.03$ & $33.48$ & $28.94$ \\
Gate + memory + row FI & $57.17$ & $53.99$ & $50.28$ & $42.54$ \\
\bottomrule
\end{tabular}
\end{table}

The real-table check reuses 12 model-independent, non-TabArena OpenML-CC18 tasks,
four fixed support/query splits per task, 409 distinct support rows, and 128
queries. Inputs are byte-identical across conditions. Because this check was
designed after observing the RBF results, it tests whether the main effects also
appear on real tables; it is not a second test specified in advance. Intervals
resample tasks while retaining all four splits.

\begin{figure}[htbp]
\centering
\includegraphics[width=0.94\textwidth]{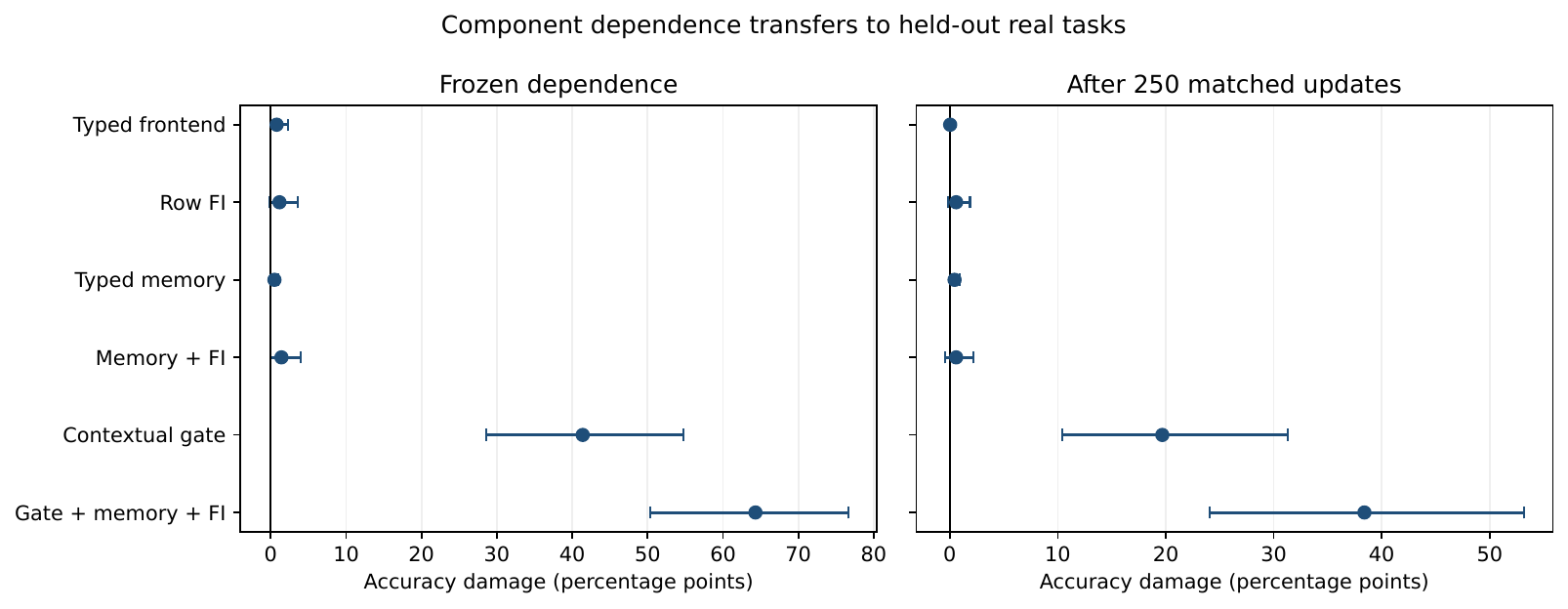}
\caption{\textbf{Component check on held-out real tables.} Points are
task-macro accuracy damage and bars are task-cluster 95\% intervals. Read task-macro damage before and after the same recovery budget. Gate
effects also appear in this exploratory real-table panel; task-cluster intervals
reflect task variation, not training-seed variation.}
\label{fig:component-recovery-real}
\end{figure}

Table~\ref{tab:component-recovery-real-datasets} gives the per-dataset effects
after matched recovery. It distinguishes broad component dependence from
effects concentrated on particular tables: gate deletion hurts all 12 datasets,
whereas row-FI deletion has its largest effect on Madelon (OpenML 1485).
The supplement provides both
recovery endpoints, split-level scores, and input hashes, with a script that
reconstructs this table and checks the means in Figure~\ref{fig:component-recovery-real}.
\begin{table}[!htbp]
\centering
\small
\setlength{\tabcolsep}{3.5pt}
\caption{\textbf{Which real datasets depend on each component?} OpenML dataset
IDs identify the 12 datasets in Figure~\ref{fig:component-recovery-real}.
Reference is the unmodified model's accuracy (\%); other columns are accuracy
losses (percentage points), averaged over four identical support/query splits.
Every model receives 250 matched recovery updates. Negative losses favor
deletion. M denotes typed memory; FI denotes row feature interaction.}
\label{tab:component-recovery-real-datasets}
\begin{tabular}{rrrrrrrr}
\toprule
Dataset ID & Reference & Frontend & Row FI & Memory & M+FI & Gate & Gate+M+FI \\
\midrule
11 & 99.02 & $-0.59$ & $+0.20$ & $+0.20$ & $+0.00$ & $+15.62$ & $+53.32$ \\
182 & 88.28 & $-0.39$ & $-0.39$ & $+0.00$ & $-0.20$ & $+19.73$ & $+68.16$ \\
1049 & 91.41 & $+0.78$ & $+0.00$ & $+0.39$ & $+0.00$ & $+9.18$ & $+18.36$ \\
1120 & 86.72 & $+0.00$ & $+0.59$ & $-0.20$ & $+0.20$ & $+8.79$ & $+14.84$ \\
1462 & 100.00 & $+0.00$ & $+0.00$ & $+0.00$ & $+0.00$ & $+2.15$ & $+14.06$ \\
1475 & 50.59 & $-0.59$ & $+0.00$ & $+0.39$ & $-0.39$ & $+34.57$ & $+37.30$ \\
1485 & 73.24 & $+1.17$ & $+7.23$ & $+2.93$ & $+8.59$ & $+14.45$ & $+23.24$ \\
1487 & 95.31 & $+0.20$ & $+0.59$ & $+0.20$ & $+0.20$ & $+9.18$ & $+12.11$ \\
1501 & 93.95 & $-0.39$ & $+0.00$ & $+0.78$ & $+0.59$ & $+70.51$ & $+75.78$ \\
1510 & 97.27 & $+0.20$ & $-0.20$ & $+0.39$ & $+0.00$ & $+1.17$ & $+5.27$ \\
40975 & 97.46 & $-0.20$ & $-0.39$ & $+0.39$ & $+0.00$ & $+13.48$ & $+79.30$ \\
40982 & 79.69 & $+0.20$ & $-0.59$ & $-0.20$ & $-1.95$ & $+37.30$ & $+58.98$ \\
\midrule
Mean & 87.74 & $+0.03$ & $+0.59$ & $+0.44$ & $+0.59$ & $+19.68$ & $+38.40$ \\
\bottomrule
\end{tabular}
\end{table}

\FloatBarrier

\section{Boundary Conditions for the Refinement Account}
\label{sec:supporting-mechanisms}

Refinement needs enough information to identify the task, and its progress
needs a task-relevant measure. We separate identifying a rule from predicting
it accurately, inspect whether sparse-evidence failures arise in later layers,
and compare representation rank with agreement to known targets. These tests
explain why larger representations or later states need not improve every
measure of task learning.

\FloatBarrier
\subsection{Does better execution imply better sparse rule identification?}

This test asks whether scaling helps every part of rule learning. The weights and
coordinates are fixed while the labeling function varies over a fixed set of
candidate rules. Sixteen rules
in each of three complexity families share 1,024 support and 2,048 query
coordinates. With 32 or 128 independent rows, repeated rows fill 1,024
support positions. The decoder selects the rule within each family whose target has
minimum cross-entropy to the predicted surface; the label-blind top-1 ceiling is
$1/16$. Balanced random labels preserve class counts and recover 3/48 rules in
total, the pooled chance count.

\begin{figure}[htbp]
\centering
\includegraphics[width=\linewidth]{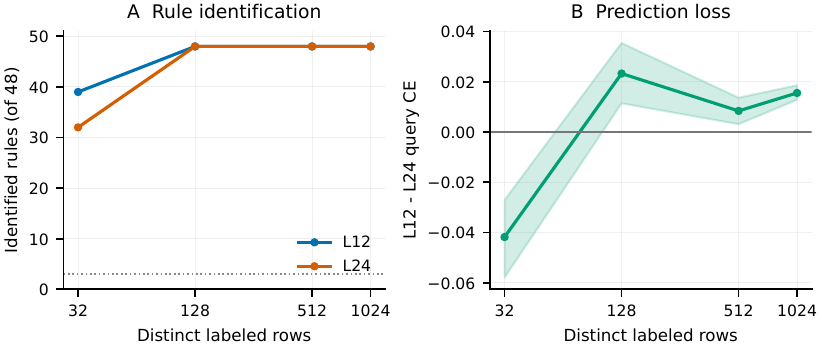}
\caption{\textbf{Rule identification and prediction accuracy improve differently.} L12
identifies more rules from 32 independent rows, while L24 achieves lower
query CE after both checkpoints identify every rule. The two panels
separate retrieval counts from the paired CE difference (positive favors L24;
95\% within-family function-bootstrap band). Executed support positions remain fixed at 1,024.}
\label{fig:program-evidence-scaling-app}
\end{figure}

Read retrieval as rule identification and own-rule query CE as prediction quality in
Figure~\ref{fig:program-evidence-scaling-app}. At 32 independent rows L12 retrieves
39/48 rules and L24 32/48. At 128 rows both reach 48/48; L24 then has lower
own-rule query CE by 0.02329, and by 0.00842 and 0.01557 at 512 and 1,024 rows.
The larger-depth checkpoint therefore does not uniformly identify sparse rules
better. Repetition holds support positions fixed but changes multiplicity; this
is a unique-evidence control, not a substitute for the distinct-prefix experiment
in Section~\ref{app:evidence-protocol}.

\FloatBarrier
\subsection{Does a rule found early disappear in later layers?}

To test whether later layers lose a rule that was already found, we reuse the
16-rule low-, medium-, and high-complexity families specified in advance, identical support
and query coordinates, 2,048 query rows, and the same 32-row repetition order as
the evidence sweep. Every layer is decoded through the shared final normalization
and head. Rule retrieval is computed within each family, so a label-blind output
has a $1/16$ ceiling. We ask whether retrieval falls by at least two rules from its
best intermediate layer to the final layer, or whether the missing rule was never
recovered at any layer.

\begin{figure}[htbp]
\centering
\includegraphics[width=\linewidth]{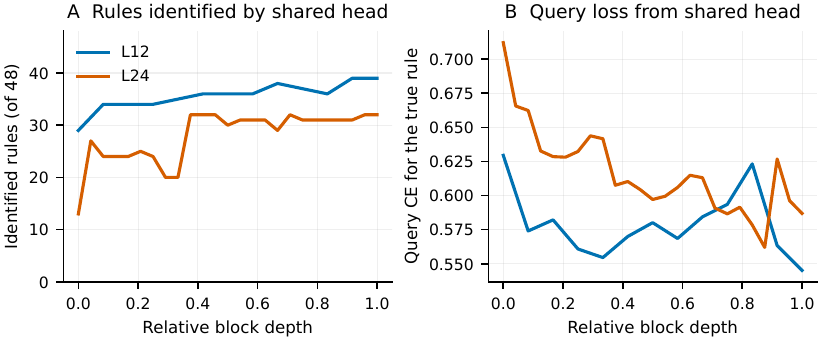}
\caption{\textbf{Layerwise readouts under sparse evidence.}
Both checkpoints see the same 32 distinct support rows repeated to 1,024 positions.
The final normalization and decoder are reused at every layer. Panel A measures
rule retrieval; panel B measures own-rule query CE. Neither curve assumes
that the shared head is an optimal intermediate decoder.}
\label{fig:sparse-program-trajectory}
\end{figure}

\FloatBarrier

The primary criterion is a drop of at least two rules from the best layer to the final layer.
L24 never exceeds its final 32/48 retrieval, while L12 ends at 39/48; the observed
layerwise result therefore rejects late loss of a rule that was already decodable.
A different intermediate readout could still recover additional information. Cross-checkpoint differences can
also reflect width, frontend and training history.

\FloatBarrier
\subsection{Can response rank be high when task performance is poor?}

The response-rank experiment fixes 128 distinct support coordinates, repeats them to
1,024 positions in one shared order, and uses 512 fixed query coordinates.
Sixteen Fourier rules in each of three families supply 48 target surfaces.
Coherent and class-count-preserving shuffled labels share the same inputs.
Hidden query responses are flattened, and spectral measures use their
$48\times48$ Gram matrix. We remove each response's spatial mean and normalize
its norm before centering across rules. Retrieval uses cross-entropy against
the 48 targets, and CKA compares response and target geometries.

\begin{figure}[htbp]
\centering
\includegraphics[width=\linewidth]{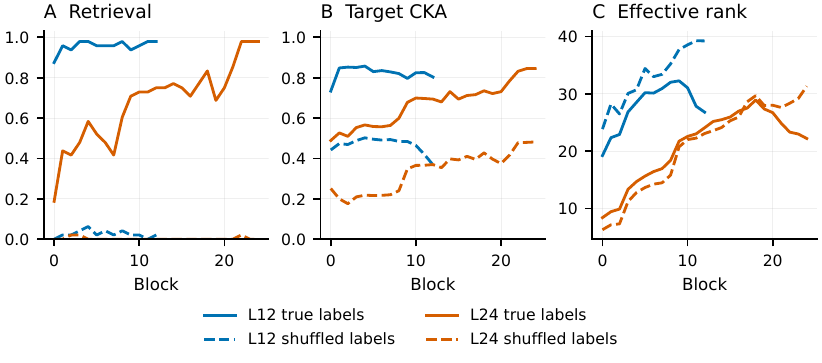}
\caption{\textbf{Agreement with the target rules, rather than raw rank, tracks
task performance.} Outputs from the true labels agree with the target-rule
structure; outputs from shuffled labels may have higher rank but fail to recover the rules.}
\label{fig:program-response-rank-app}
\end{figure}

Figure~\ref{fig:program-response-rank-app} should be read jointly: final outputs
from the true labels retrieve 47/48 rules for both checkpoints, whereas shuffled L12 and
L24 retrieve 1/48 and 0/48 despite sometimes higher effective rank. Thus raw rank
alone does not measure task performance, which motivates the controlled rank limit in
Section~\ref{sec:width-dose-table}. This 512-query, 48-way retrieval probe differs
from the 2,048-query within-family rule set, so its counts are not pooled.

\FloatBarrier
\subsection{Does a layerwise readout approach a known prediction target?}

As a complementary target-space check, we generate 64 binary logistic episodes
whose conditional distribution is known exactly. Applying the shared final
normalization and decoder at every L24 block gives a layerwise predictor
$p_{e,\ell}$. Its Bayes excess cross-entropy falls from 0.13047 at block 0 to
0.02559 at block 12 and 0.01476 at block 24. Blocks 13--24 therefore remove
another 0.01082 excess cross-entropy (paired episode-bootstrap 95\% interval
[0.00858, 0.01311]). A support-only logistic ERM reaches 0.01416 and serves only
as a finite-sample reference. This known-generator result agrees with the
class-prototype intervention in the main text while relying on a narrower binary
task family.

The solver probe uses 64 paired episodes, 256 support rows, 512 query rows, and
feature dimension in $\{2,4,8\}$. Every state is decoded through the final
normalization and head; final decoded logits reproduce production logits within
$10^{-3}$. Affine-logit readout $R^2$ is 0.97745 for L12 and 0.98770 for L24.

Intervals resample episodes. Cross-checkpoint differences are not depth-only, and
the candidate optimizer directions are too similar to identify a unique internal
optimizer. This supports later-block progress toward a known target, without identifying
a unique internal optimization algorithm.

\clearpage
\section{Scope, Limitations, and Reproducibility}
\label{app:limitations}

\paragraph{Analytic and learned updates.}
The proof concerns a locally smooth leave-one-out support objective with at least
two examples per class. The kernel approximation requires small
$\beta\|K\|_2/\alpha$; neither its exact gradient nor its scale is
computed explicitly by the network. Query risk and repeated learned-block descent
are empirical questions, not consequences of the proposition.

\paragraph{Scope of the interventions.}
Mechanism tests emphasize synthetic classification with known rules. The real-table
recovery panel tests a narrower transfer claim. A shared final decoder can miss
intermediate information, and disabling components at inference can create states
not encountered during training. Short recovery does not replace training all
ablated architectures to convergence. The fixed-budget FFN grid has one training
seed; paired episode intervals do not quantify training variance.

\paragraph{What scale comparisons establish.}
L12 and L24 differ in depth, width, and training history. The change in their
accuracy gap with context size therefore applies only to this pair of checkpoints.
The rank-32 projection measures sensitivity inside one trained model. The sparse-rule deficit and intermediate reversals
rule out uniform improvement with additional depth or context. Parameter ratios
against TabFM are storage comparisons; they do not establish a matched-compute
training advantage.

\paragraph{Benchmark and release scope.}
Section~\ref{app:benchmark-provenance} records which headline results were
reaggregated from task or fold outputs and identifies the distinct L24
checkpoints used by AMLB29 and TabArena. The TabArena ranking belongs to the
fixed 38-dataset, fold-0 snapshot and its
heterogeneous reference budgets. Overlapping OpenML views and metrics are not
independent trials, and small mean differences alone do not establish statistical
superiority. Complete RefineICL TALENT evaluations are unavailable in the reported
evidence and do not enter a comparison. Regression is outside scope.
The code supplement contains the model, synthetic priors, training
loop, optimizer, and JSON recipes, plus focused tests and a CPU smoke run.
It documents the benchmark-checkpoint mapping and training stages. Trained
weights, benchmark data, and the real-informed continuation manifests are not
included; the supplied source therefore supports model construction and
synthetic training rather than a complete replay of the TabArena continuation.

\end{document}